\documentclass[lettersize,journal]{IEEEtran}
\usepackage{amsmath,amsfonts}
\usepackage{algorithmic}
\usepackage{algorithm}
\usepackage{array}
\usepackage[caption=false,font=normalsize,labelfont=sf,textfont=sf]{subfig}
\usepackage{textcomp}
\usepackage{stfloats}
\usepackage{url}
\usepackage{verbatim}
\usepackage{graphicx}
\usepackage{cite}
\usepackage{booktabs}
\usepackage[margin=1in]{geometry}
\usepackage{xcolor}
\usepackage{hyperref}
\usepackage{epigraph}

\usepackage{amsthm}
\usepackage{tabularx}
\usepackage{booktabs} 
\usepackage{etoolbox}
\usepackage{makecell}
\theoremstyle{plain}
\newtheorem{theorem}{Theorem}[section]

\theoremstyle{definition}
\newtheorem{definition}[theorem]{Definition}

\theoremstyle{remark}

\usepackage{amssymb}
\usepackage{mydef}
\usepackage{bm}
\begin{document}

\title{RiemannGL: Riemannian Geometry Changes Graph Deep Learning}

\author{Li Sun, Qiqi Wan, Suyang Zhou, Zhenhao Huang, Philip S. Yu,~\IEEEmembership{Fellow,~IEEE}
    
\IEEEcompsocitemizethanks{
\IEEEcompsocthanksitem Li Sun is with Beijing University of Posts and Telecommunications, Beijing 100876, China. E-mail: lsun@bupt.edu.cn.
\IEEEcompsocthanksitem Qiqi Wan is with Beihang University, Beijing 100191, China. E-mail: wanqiq@buaa.edu.cn.
\IEEEcompsocthanksitem Suyang Zhou is with East China Normal University, Shanghai 200062, China. E-mail: 51285902056@stu.ecnu.edu.cn.
\IEEEcompsocthanksitem Zhenhao Huang is with North China Electric Power University, Beijing 102206, China. E-mail: huangzhenhao@ncepu.edu.cn.
\IEEEcompsocthanksitem Philip S. Yu is with the Department of Computer Science, University of Illinois Chicago, Chicago IL 60607, USA. E-mail: psyu@uic.edu.
}
 
}

\markboth{RiemannGL: Riemannian Geometry Changes Graph Deep Learning}%
{Sun \MakeLowercase{\textit{et al.}}: RiemannGL: Riemannian Geometry Changes Graph Deep Learning}

\maketitle

\begin{abstract}
Graphs are ubiquitous, and learning on graphs has become a cornerstone in artificial intelligence and data mining communities.  
Unlike pixel grids in images or sequential structures in language, graphs exhibit a typical non-Euclidean structure with complex interactions among the objects.
This paper argues that Riemannian geometry provides a principled and necessary foundation for graph representation learning, and that Riemannian graph learning should be viewed as a unifying paradigm rather than a collection of isolated techniques.
While recent studies have explored the integration of graph learning and Riemannian geometry, most existing approaches are limited to a narrow class of manifolds, particularly hyperbolic spaces, and often adopt extrinsic manifold formulations.
We contend that the central mission of Riemannian graph learning is to endow graph neural networks with intrinsic manifold structures, which remains underexplored.
To advance this perspective, we identify key conceptual and methodological gaps in existing approaches and outline a structured research agenda along three dimensions: manifold type, neural architecture, and learning paradigm.
We further discuss open challenges, theoretical foundations, and promising directions that are critical for unlocking the full potential of Riemannian graph learning.
This paper aims to provide a coherent viewpoint and to stimulate broader exploration of Riemannian geometry as a foundational framework for future graph learning research.
\end{abstract}

\begin{IEEEkeywords}
Graph Learning, Riemannian Graph Learning, Non-Euclidean Geometry, Taxonomy, Foundation Model, AI for Science.
\end{IEEEkeywords}

\tikzstyle{root}=[
    draw=mygreen!60,                            
    rounded corners,                       
    minimum height=1em,                    
    fill=mygreen!60,                          
    text opacity=1,                        
    align=center,                          
    text=black,                            
    font=\scriptsize,                      
    inner xsep=4pt,                        
    inner ysep=4pt                         
]
\tikzstyle{second_one}=[
    draw=secondone!60,
    rounded corners,
    minimum height=1em,
    text opacity=1,
    fill=secondone!60,
    align=center,
    text=black,
    font=\scriptsize,
    inner xsep=4pt,
    inner ysep=4pt,
]
\tikzstyle{second_two}=[
    draw=secondtwo!60,
    rounded corners,
    minimum height=1em,
    text opacity=1,
    fill=secondtwo!60,
    align=center,
    text=black,
    font=\scriptsize,
    inner xsep=4pt,
    inner ysep=4pt,
]
\tikzstyle{second_three}=[
    draw=secondthree!60,
    rounded corners,
    minimum height=1em,
    text opacity=1,
    fill=secondthree!60,
    align=center,
    text=black,
    font=\scriptsize,
    inner xsep=4pt,
    inner ysep=4pt,
]
\tikzstyle{second_four}=[
    draw=secondfour!60,
    rounded corners,
    minimum height=1em,
    text opacity=1,
    fill=secondfour!60,
    align=center,
    text=black,
    font=\scriptsize,
    inner xsep=4pt,
    inner ysep=4pt,
]
\tikzstyle{third_one}=[draw=secondone,
    rounded corners,minimum height=1em,
    fill=secondone!30,text opacity=1, align=center,text=black,align=left,font=\scriptsize,
    inner xsep=3pt,
    inner ysep=1pt,
]
\tikzstyle{third_two}=[draw=secondtwo,
    rounded corners,minimum height=1em,
    fill=secondtwo!30,text opacity=1, align=center,text=black,align=left,font=\scriptsize,
    inner xsep=3pt,
    inner ysep=1pt,
]
\tikzstyle{third_three}=[draw=secondthree,
    rounded corners,minimum height=1em,
    fill=secondthree!30,text opacity=1, align=center,text=black,align=left,font=\scriptsize,
    inner xsep=3pt,
    inner ysep=1pt,
]
\tikzstyle{third_four}=[draw=secondfour,
    rounded corners,minimum height=1em,
    fill=secondfour!30,text opacity=1, align=center,text=black,align=left,font=\scriptsize,
    inner xsep=3pt,
    inner ysep=1pt,
]

\tikzstyle{nodefeatureleaf_one}=[
    draw=thirdone,
    rounded corners,
    minimum height=1em,
    fill=thirdone,
    text opacity=1, 
    align=center,
    text=black,
    align=left,font=\scriptsize,
    inner xsep=3pt,
    inner ysep=1pt,
]
\tikzstyle{nodefeatureleaf_two}=[
    draw=thirdtwo,
    rounded corners,
    minimum height=1em,
    fill=thirdtwo,
    text opacity=1, 
    align=center,
    text=black,
    align=left,font=\scriptsize,
    inner xsep=3pt,
    inner ysep=1pt,
]
\tikzstyle{nodefeatureleaf_three}=[
    draw=thirdthree,
    rounded corners,
    minimum height=1em,
    fill=thirdthree,
    text opacity=1, 
    align=center,
    text=black,
    align=left,font=\scriptsize,
    inner xsep=3pt,
    inner ysep=1pt,
]
\tikzstyle{nodefeatureleaf_four}=[
    draw=thirdfour,
    rounded corners,
    minimum height=1em,
    fill=thirdfour,
    text opacity=1, 
    align=center,
    text=black,
    align=left,font=\scriptsize,
    inner xsep=3pt,
    inner ysep=1pt,
]

\begin{figure*}[ht]
\centering
\resizebox{\textwidth}{!}{
\begin{forest}
  for tree={
  forked edges,
  grow=east,
  reversed=true,
  anchor=base west,
  parent anchor=east,
  child anchor=west,
  base=middle,
  font=\scriptsize,
  rectangle,
  line width=0.7pt,
  draw=black,
  rounded corners,
  align=left,
  minimum width=2em,
  s sep=3pt,
  l sep=9pt,
  inner xsep=3pt,
  inner ysep=1pt,
  },
  where level=1{text width=4.5em}{},
  where level=2{text width=4em,font=\scriptsize}{},
  where level=3{font=\scriptsize}{},
  where level=4{font=\scriptsize}{},
  where level=5{font=\scriptsize}{},
  [Riemannian Graph Learning, root, rotate=90, anchor=north, edge=black
    [Manifold Type, second_one, edge=black,text width=6em,
        [Hyperbolic Manifold, third_one, text width=6em, edge=black
            [{ HGNN\cite{hnnnips19}, 
    p-VAE\cite{mathieu2019continuous},
    ROTE\cite{ATTH20},
    HTF\cite{HTF},
    HGCF\cite{sun2021hgcf},
    DT-GCN\cite{DT-GCN21},
    ACE-HGNN\cite{ace-hgnn2021},
    CurvGAN\cite{li2022curvature},\\
    HIE\cite{hie23icml},
    HDD\cite{LinCMT23},
    HGWaveNet\cite{bai2023hgwavenet},
    HyperIMBA\cite{HyperIMBA23},
    SHAN\cite{li2023multi}, 
    FFHR\cite{FFHR},
    H-Diffu\cite{FengZFFWLS23},\\
    HypMix\cite{HypMix24},
    HMPTGN\cite{LeT24},
    MSGAT\cite{msgat2024icdm}, 
    HGCH\cite{HGCH24},
    THGNets\cite{liu2025thgnets}, 
    HGDM\cite{HGDM24AAAI},\\
    HyperDefender\cite{MalikGK25},
    HyperGCL\cite{HyperGCL24kdd},
    HEDML\cite{HEDML25www},
    HGCN\cite{hgcn2019nips}, 
  $\mathcal{TC}$-flow \cite{BoseSLPH20}, 
  WHC\cite{BoseSLPH20},\\
  HVGNN\cite{tgnn2021}, 
  LGCN\cite{LGCN21WWW},
  H2H-GCN\cite{h2hgcn2021},
  RotDif\cite{QiaoFLLH0Y23},
  R-HGCN\cite{RHGCNXue2024AAAI}, 
  Hypformer\cite{yang2024hypformer},\\
  LorentzKG\cite{LorentzKG}, 
  LSEnet\cite{sunLSEnet}, 
  HypDiff\cite{HypDiff24},
  MHR\cite{feng2025mhr},
  Hgformer\cite{LHGCN},
  HVGAE\cite{liu2025hyperbolic}}, nodefeatureleaf_one, text width=29em, edge=black 
            ]
        ]
        [Spherical Manifold, third_one, text width=6em, edge=black 
            [{
            DeePSphere \cite{defferrard2020deepsphere} SphereNet \cite{liu2021spherical} SMP\cite{smp22iclr} LHML\cite{guo2022learning}},
            nodefeatureleaf_one, text width=29em, edge=black 
            ]
        ]
        [Constant Curvature Space, third_one, text width=8em, edge=black 
            [{DyERNIE\cite{DyERNIE},
    MVAE\cite{skopek2019mixed},
    M2GNN\cite{wang2021mixed},
    SelfMGNN\cite{sun2022aaai},
    HGE\cite{pan2024hge},
    IME\cite{wang2024ime},
    MCKGC\cite{gao2025mixed}}, nodefeatureleaf_one, text width=27em, edge=black 
            ]
        ]
        [Product and Quotient Space, third_one, text width=9em, edge=black 
            [{$\kappa$-GCN\cite{Bachmann2020ICML}, 
    DyERNIE\cite{DyERNIE},
    MAVE\cite{skopek2019mixed},
    M2GNN\cite{wang2021mixed},
    SelfMGNN\cite{sun2022aaai},
    HGE\cite{pan2024hge}, \\
    ProGDM\cite{ProGDM24},  
    IME\cite{wang2024ime},
    GeoMancer\cite{Geomancer25gao},
    GraphMoRE\cite{GraphMoRE},
    MCKGC\cite{gao2025mixed},\\
    RiemannGFM\cite{RGFM25sun} }, nodefeatureleaf_one, text width=26em, edge=black 
            ]
        ]
        [Pseudo-Riemannian Manifold, third_one, text width=9em, edge=black 
            [{
            PseudoNet\cite{law2020ultrahyperbolic}, UltraNet\cite{law2021ultrahyperbolic}, Directed-Pseudo\cite{sim2021directed}, $\mathcal{Q}$-GCN\cite{Xiong2022NeurIPS}, UltraE\cite{xiong2022ultrahyperbolic},  
    $\mathcal{Q}$-GT\cite{qgt25}}, nodefeatureleaf_one, text width=26em, edge=black 
            ]
        ]
        [Grassmann Manifold, third_one, text width=6em, edge=black 
            [{EGG\cite{EGG24zhou}, GyroGr\cite{GyroGr}, SPSD\cite{GyroSpsd++24son}, CAGM\cite{ma2024cross}, sLGm\cite{wu2024multiple}, CLRSR\cite{wang2018cascaded}, G-ALDNLR\cite{piao2019double}, \\PGM-HLE\cite{batalo2022analysis}, GrCNF\cite{yataka2023grassmann}, GRLGQ\cite{mohammadi2024generalized}, AMCGM\cite{wu2022attention} }, nodefeatureleaf_one, text width=29em, edge=black 
            ]
        ]
        [SPD Manifold, third_one, text width=5em, edge=black 
            [{
            Graph-CSPNet\cite{ju2024graph}, GyroGr\cite{nguyen2023building}, SPSD\cite{GyroSpsd++24son}, GBWBN\cite{wang2025learning}, SPDNet\cite{huang2016riemannian}, DMTNet\cite{zhang2017deep}, \\ SPDNetNAS\cite{sukthanker2021neural}, SymNet\cite{wang2022symnet}, MSNet\cite{chen2023riemannian} 
            }, nodefeatureleaf_one, text width=30em, edge=black 
            ]
        ]
        [Generic Manifold , third_one, text width=5em, edge=black 
            [{
            RicciNet\cite{sunRiccinet}, \textsc{Congregate}\cite{sunCONGREGATE}, R-ODE\cite{sun2024r}, MSG\cite{MSG2024nips}, MofitRGC\cite{sun2024aaai}, Pioneer\cite{Pioneer2025AAAI}, \\CurvGAD\cite{CurvGAD}, GBN \cite{gbn25sun} 
            }, nodefeatureleaf_one, text width=30em, edge=black 
            ]
        ]
    ]
    [Neural Architectures, second_two, edge=black,text width=6em
        [Riemannian Convolutional Network, third_two, text width=11em, edge=black 
            [{HGCN \cite{hgcn2019nips} $\kappa$-GCN \cite{Bachmann2020ICML} HAT \cite{HGAT2021} LGCN \cite{LGCN21WWW} H2H-GCN \cite{h2hgcn2021} TGNN \cite{tgnn2021} \\ACE-HGNN \cite{ace-hgnn2021} DHGAN \cite{HGAT2022CIKM} SelfMGNN \cite{sun2022aaai} CGNN \cite{cgnn2022} $\mathcal{Q}$-GCN \cite{Xiong2022NeurIPS} \\DH-GAT \cite{HGAT2023SIGIR} $\mathcal{P}$-HGCN \cite{RHGCNXue2024AAAI} MSGAT \cite{msgat2024icdm} MSG \cite{MSG2024nips} D-GCN \cite{sun2024aaai}}, nodefeatureleaf_two, text width=22em, edge=black 
            ]
        ]
        [Riemannian Variational Auto Encoder, third_two, text width=11em, edge=black 
            [{$\mathcal{P}$-VAE \cite{mathieu2019continuous} Mixed-curvature VAE \cite{skopek2019mixed} GM-VAE \cite{cho2023hyperbolic} HVGAE \cite{liu2025hyperbolic}}, nodefeatureleaf_two, text width=22em, edge=black 
            ]
        ]
        [Riemannian Transformer, third_two, text width=8em, edge=black 
            [{
            Hypformer \cite{yang2024hypformer} LHGCN\cite{LHGCN}}, nodefeatureleaf_two, text width=10em, edge=black 
            ]
        ]
        [Riemannian Neural ODE, third_two, text width=8em, edge=black 
            [{R-ode \cite{sun2024r} Pioneer \cite{Pioneer2025AAAI} $\mathcal{TC}$-flow\cite{BoseSLPH20} MCNFs\cite{MCNFs2020NeurIPS} CNFM\cite{Ben-HamuCBANGCL22} CIDER\cite{LouXFE23}}, nodefeatureleaf_two, text width=24em, edge=black 
            ]
        ]
        [Riemannian Denoising Diffusion and SDE, third_two, text width=13em, edge=black 
            [{GDSS\cite{jo2022score} RDM\cite{HuangABPC22} SDDM\cite{sun2023sddm} LogBM\cite{icmlJoH24}\\ CFG++\cite{ChungKPNY25} HGCN2SP\cite{WuZLC24} MPDR\cite{YoonJN023}}, nodefeatureleaf_two, text width=16em, edge=black 
            ]
        ]
        [Riemannian Flow Model and Flow Matching, third_two, text width=13em, edge=black 
            [{RFM\cite{flowmatching} FlowMM\cite{flowmm24} M-FFF\cite{SorrensonDRHK24} CNFM\cite{Ben-HamuCBANGCL22}}, nodefeatureleaf_two, text width=16em, edge=black 
            ]
        ]
    ]
    [Learning Paradigms, second_four, edge=black,text width=6em
           [Unsupervised learning, third_four, text width=7em, edge=black 
                [{HypMix\cite{HypMix24} RicciNet\cite{sunRiccinet} \textsc{Congregate}\cite{sunCONGREGATE} LSEnet\cite{sunLSEnet}}, nodefeatureleaf_four, text width=20em, edge=black 
                ]
        ]
        [Semi-supervised Learning, third_four, text width=8em, edge=black 
                [{HGCN\cite{hgcn2019nips} THGNets\cite{liu2025thgnets} HyperDefender\cite{MalikGK25} R-HGCN\cite{RHGCNXue2024AAAI} HGWaveNet\cite{bai2023hgwavenet} HyperIMBA\cite{HyperIMBA23}\\$\kappa$-GCN \cite{Bachmann2020ICML} $\mathcal{Q}$-GCN \cite{Xiong2022NeurIPS} UltraNet\cite{law2020ultrahyperbolic} Pioneer\cite{Pioneer2025AAAI} MSG\cite{MSG2024nips} GraphMoRE\cite{GraphMoRE}}, nodefeatureleaf_four, text width=26em, edge=black 
                ]
        ]
        [Self-supervised Learning, third_four, text width=8em, edge=black 
                [{DyERNIE\cite{DyERNIE} MofitRGC\cite{sun2024aaai} HGE\cite{pan2024hge} SHAN\cite{li2023multi} RieGrace\cite{sun2023self}}, nodefeatureleaf_four, text width=20em, edge=black 
                ]
        ]
        [Foundation Model and Transfer Learning, third_four, text width=12em, edge=black 
                [{RiemannGFM\cite{RGFM25sun} GFT\cite{GFT}}, nodefeatureleaf_four, text width=10em, edge=black 
                ]
        ]
    ]
  ]
\end{forest}%
} 
\caption{Taxonomy of Riemannian Graph Learning.}
\label{pttaxonomy}
\end{figure*}
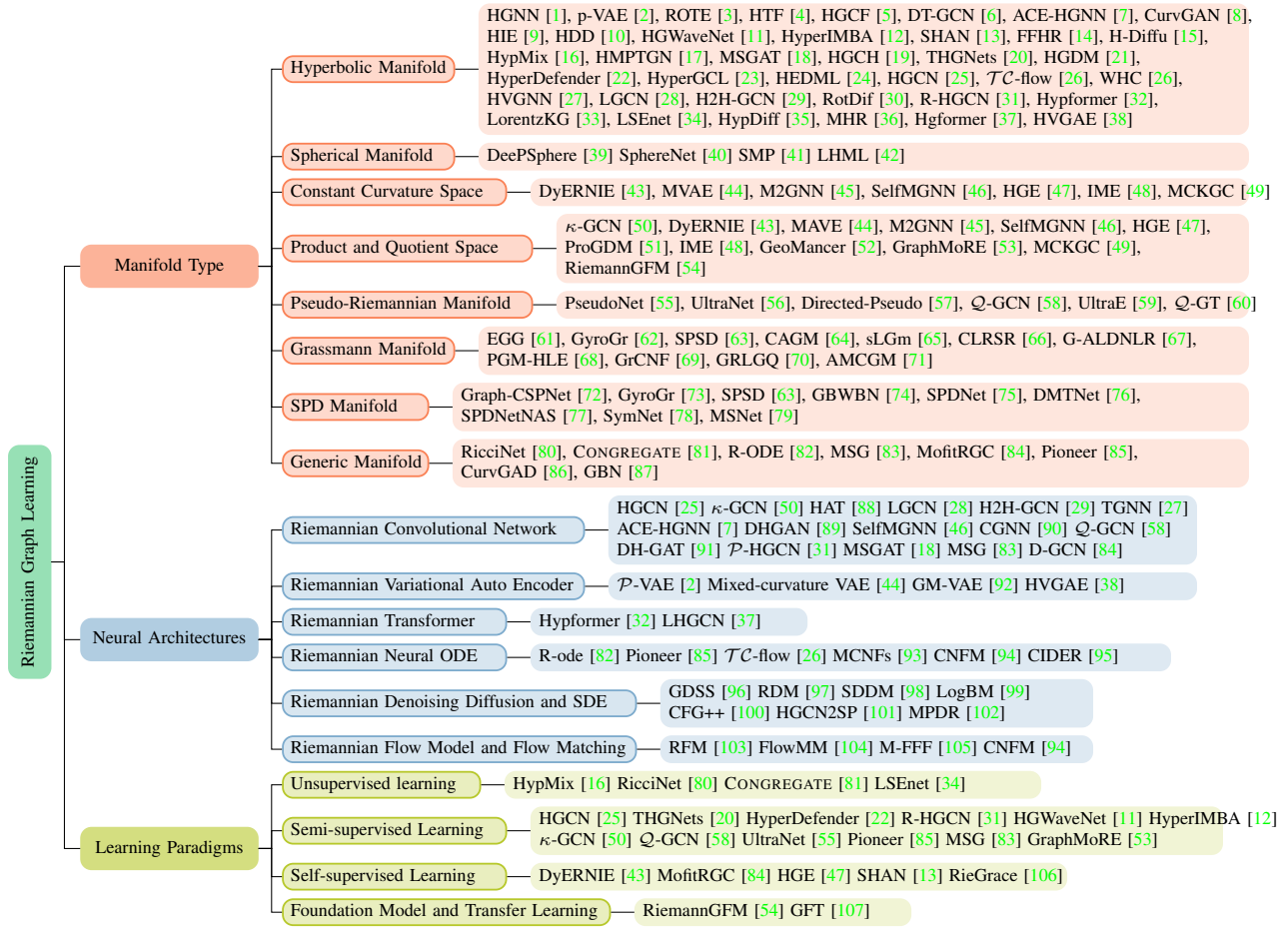

{\section{Introduction}}

\begin{flushright}
\emph{“Manifolds are the language of the cosmos; curvature is the cipher of spacetime.”}
\end{flushright}

\IEEEPARstart{T}{he} graph is a non-Euclidean structure that naturally describes correlations among objects, and it has been extensively studied in artificial intelligence and data mining communities. Graphs are omnipresent and routinely find themselves in recommender systems \cite{wu22Recommender}, social media analysis \cite{Hosseini23social}, transportation networks \cite{Rahmani23Transportation}, financial technology \cite{XiangCSZL22}, biochemical moleculars \cite{godwin2021simple}, physical interaction systems \cite{BishnoiBJRK23}, etc. For instance, describing a collection of financial transactions as a graph, each node refers to an individual account, while edges are utilized to represent the transactions among accounts. For a molecular, nodes are the atoms and edges signify the biochemical interaction between them. Discovering knowledge from graphs is thus beneficial for society, science, and humanity. Deep learning excels in data representation and pattern recognition, while learning on graphs is challenging. 
Graphs exhibit a typical non-Euclidean pattern of complex structures, unlike pixel grids in images or the sequential structure in language , which are more amenable to traditional deep learning methodologies. Graph convolution network \cite{gcn} marked an initial success in graph deep learning, and over the last decade the field has witnessed substantial progress, ranging from graph variational autoencoder \cite{vgae} to graph transformer \cite{graphtransformer}. The vast majority of graph deep models pay little attention to the choice of representation spaces and consider representations in Euclidean space. This fundamental mismatch renders inadequate expressiveness in modeling the rich structural complexity.

Riemannian geometry is a mature scientific discipline that systematically studies metric structures and curvature-related phenomena. 
In the history of science, Riemannian geometry shows its fundamental importance in Einstein’s theory of relativity (formulated on pseudo-Riemannian manifolds), Hamiltonian mechanics (working with a generic manifold), and still plays a crucial role in quantum physics and computational biology \cite{hamQuantum, hambiology}. 
With over a century of development, Riemannian geometry has been firmly established. It provides the concept of Riemannian manifolds as a principled framework for modeling geometric structures, grounded in an elegant framework of differential and algebraic geometry. Beyond basic notions of distance and angles, it offers clean, closed formulations to measure graph similarity \cite{sun2022self}, to preserve 3D equivariance \cite{EDM}, and to reveal cluster boundaries \cite{sun2023deepricci}. Moreover, Riemannian geometry provides geometric tools to facilitate graph generation \cite{tang2021generalizeda}, to control physical characteristics \cite{Pioneer2025AAAI}, to solve partial differential equations \cite{yue2025hyperbolic-pdeGNN}, and to describe structural evolution \cite{sunLSEnet}.

It is not until recent years that Riemannian geometry has been combined with learning on graphs. Among Riemannian manifolds, the hyperbolic space is first recognized in the graph domain, and a series of hyperbolic graph convolution networks have been designed and reported the superior expressiveness to their Euclidean counterparts for modeling hierarchical graphs \cite{hgcn2019nips}. The hyperbolic space also naturally expresses the graph organization pattern, such as hierarchical clustering \cite{sunLSEnet}. Spherical space is well aligned with cyclical structures such as cliques \cite{riemannianbook}, and offers a geometric prior for the Earth’s interacting system \cite{Pioneer2025AAAI}. Grassmann manifold, a representative of matrix manifolds, finds its role in studying orthogonality and low-rank optimization \cite{wang2018cascaded}. Modeling directed graphs can leverage Einstein’s spacetime, which is a kind of pseudo-Riemannian manifold \cite{law2023spacetime}. The product manifold construction has been shown to benefit graph contrastive learning \cite{RGFM25sun}, while the concept of curvature can be harnessed to perform node clustering \cite{sunRiccinet} and to mitigate the issue of oversquashing (e.g., curvature-based graph rewriting \cite{sdrf}). In addition, residual connections \cite{katsman2023riemannian}, transformers \cite{LHGCN}, and stochastic differential equations \cite{HuangABPC22} are extended to Riemannian manifolds. Graph learning with Riemannian geometry is progressing swiftly.

It is timely to review the development trajectory of this field and identify emerging directions. 
\textbf{Riemannian Graph Learning} generally refers to the graph learning methodologies that involve Riemannian manifolds. The theoretical foundation is Riemannian geometry, which offers a rigorous mathematical framework to explore graph structure. The core idea is to connect graph structures to manifolds, so that the geometric tools are activated for graph deep learning. 
While achieving encouraging results, Riemannian graph learning remains in its infancy. 
On the one hand, previous efforts overemphasize modeling graphs in hyperbolic space. 
While hyperbolic spaces are suitable for studying hierarchies, graphs are far more complex than hierarchies, and there exists a wide spectrum of manifolds with rich geometric advantages, e.g., conformal, equivariant, quotient, and symmetric manifolds. 
The expressive power of Riemannian graph learning has yet to be fully unleashed.  
On the other hand, previous studies primarily focus on generalizing existing formulations to manifolds, and thus, the underlying mechanisms are largely inherited from the original Euclidean counterparts. However, the mission of Riemannian graph learning should be to go beyond Euclidean limitations by exploiting uniquely Riemannian geometric concepts and tools. Our perspective is to endow neural networks with intrinsic manifold properties, which are difficult to achieve even with careful design in the standard Euclidean geometry. 

\textbf{Position.} 
In this paper, we argue that Riemannian geometry should be viewed not merely as an auxiliary tool but as a foundational framework for graph learning, particularly for the development of graph foundation models.
We contend that the intrinsic manifold character, rather than extrinsic formulations, should guide the design of neural architectures and learning paradigms.
We highlight that Riemannian graph learning holds significant potential for artificial intelligence for science (AI4S) and foundation models.
Since the development of science is accompanied by Riemannian geometry, as mentioned above, a geometric prior can fundamentally benefit scientific reasoning and inference. Riemannian geometry offers a lens to explore structural commonality and invariance, enabling the construction of a graph foundation model.

To articulate this position, we propose a conceptual framework that organizes Riemannian graph learning along three dimensions: manifold type, neural architecture, and learning paradigm. Eight representative manifolds are considered, including hyperbolic, spherical, constant curvature, product, quotient, pseudo-Riemannian, Grassmann, and generic manifolds. Six neural architectures are reviewed in the manifold context, including graph convolution networks, graph variational autoencoders, transformers, graph ODEs, denoising diffusion and SDEs, and flow matching. We further discuss implications for unsupervised, semi-supervised, self-supervised, and transfer learning, as well as for the construction of graph foundation models.

Rather than providing a comprehensive survey, our goal is to highlight open problems and emerging directions, and to establish a research agenda for Riemannian graph learning. We emphasize areas where existing Euclidean-based generalizations are insufficient and where intrinsic Riemannian concepts can offer unique advantages in expressive power, scalability, and interpretability. By framing the field in this way, we aim to guide future research and encourage the community to advance Riemannian approaches for graph foundation models and AI for science.

%
\textbf{Contribution Highlights.}
 Notable contributions of our paper are summarized as follows, 
 \begin{itemize}
    \item \textbf{Conceptual Framework.} We define Riemannian graph learning, review representative manifolds, and introduce a three-dimensional framework: manifold type, neural architecture, and learning paradigm. This framework serves as a lens to understand the potential and limitations of current approaches.
    \item \textbf{Guiding Insights.} We provide key insights into Riemannian graph learning. For each taxonomic category, we establish the mathematical framework, highlight important methodological advances, and offer comparative analysis and discussion.
     \item \textbf{Abundant Resources.} We collect abundant resources on Riemannian graph learning, including benchmark datasets and open-source codes, and summarize practical applications in various domains. 
     \item \textbf{Future Directions.} We identify the open issues in benchmarking, graph diversity, model expressiveness, architecture, interpretability, and scalability, while outlining their potential in graph foundation models and artificial intelligence for science.
 \end{itemize}

\noindent\textbf{Organization of this Paper.} 
The remainder of this paper is organized as follows. 
Section \ref{sec: RGL} specifies the concept of Riemannian Graph Learning, and its difference and connection to related topics. 
Section \ref{sec: preliminaries} lists the main notations and reviews the fundamentals of Riemannian geometry. 
Section \ref{sec: Taxonomies} outlines the categorization of Riemannian graph learning in terms of manifold type, neural architecture, and learning paradigm. 
Sections \ref{sec: manifold}, \ref{sec: architectures}, and \ref{sec: paradigms} comprehensively introduce the advances in Riemannian graph learning under the new taxonomy in this paper. 
Section \ref{sec: datasets} provides the datasets and open resources. 
Section \ref{sec: applications} presents a collection of practical applications across various domains. 
Section \ref{sec: future} discusses the open issues and future directions. 
We conclude this paper in Section \ref{sec: Conclusion}.

\section{Riemannian Graph Learning} \label{sec: RGL}

\subsection{Graphs and Learning on Graphs}
Graphs naturally model pairwise relationships between entities, capturing complex structural information. We formally define the core graph concepts used in this survey below. 

\begin{definition}[{Graphs}]
Let $G = (V, E)$ be a graph with a node set $V$ and an edge set $E$, where $n=|V|$ denotes the number of nodes. An edge connecting $v_i$ and $v_j$ is denoted as $e_{ij} \in E$. The topological structure is encoded by an adjacency matrix $\mathbf{A} \in \mathbb R^{n \times n}$, where $A_{ij} = 1$ if $e_{ij} \in E$, and 0 otherwise. Nodes are typically associated with a feature matrix $\mathbf{X} \in \mathbb{R}^{n \times d}$, where $d$ is the feature dimension.
\end{definition}

\begin{definition}[{Directed/Undirected Graphs}] 
In a directed graph, edges have a direction, pointing from a source node $v_i$ to a target node $v_j$, denoted as $e_{i\rightarrow j}$. Conversely, undirected graphs consist of bidirectional interactions, resulting in a symmetric adjacency matrix, i.e. $\mathbf A =\mathbf A ^\top$.
\end{definition}

\begin{definition}[{Dynamic Graphs}]
Dynamic graphs describe structural evolution over time and are typically modeled in two forms: (1) A sequence of graphs  \(G = (G^0, G^1, \dots, G^T)\) describing structure at specific time intervals;  (2) A set of timestamped interactions $\{(v_i, v_j, t_{ij})\}$, capturing real-time structural changes. 
\end{definition}

\begin{definition}[{Homophilic/Heterophilic Graphs}]
Based on the node label set $\mathcal{Y}$, homophily measures the tendency of nodes to connect with similar neighbors. This is quantified by node homophily $\mathcal{H}_{\text{node}}$ and edge homophily $\mathcal{H}_{\text{edge}}$, 
\begin{equation}
    \mathcal{H}_{\text{node}} = \frac{1}{|\mathcal{V}|} \sum_{v \in \mathcal{V}} \frac{|\{u \in \mathcal{N}(v) : y_v = y_u\}|}{|\mathcal{N}(v)|},
\end{equation}
\begin{equation}
    \mathcal{H}_{\text{edge}} = \frac{|\{(v, u) \in E : y_v = y_u\}|}{|E|},
\end{equation}
where $\mathcal{N}$  denotes the neighbors of the center node.
High values of $\mathcal{H}$ indicate strong similarity among connected nodes (homophily), whereas low values characterize heterophilic graphs, where edges typically link dissimilar entities. 
\end{definition}

\begin{definition}[{Homogeneous/Heterogeneous Graphs}]
Graphs are categorized by their node type set  $T^v$ and edge type set $T^e$. A homogeneous graph consists of a single type of node and edge (\(|{T}^v|\)=\(|{T}^e|\)=1). Conversely, a heterogeneous graph involves multiple types of entities or relations, formally defined as  \(|{T}^v| + |{T}^e| > 2\).
 \end{definition}

\vspace{-7mm}

\subsection{What is Riemannian Graph Learning?}
Unlike pixel grids in the images and sequences in the language, graphs are typical non-Euclidean data with complex structures.
The majority of traditional graph models ignore the  underlying geometric prior, and tend to result in  suboptimal performance.
Theoretically, a graph is a discrete analogy to a certain Riemannian manifold in Riemannian geometry \cite{petersenRiemannianGeometry2016}.
Empirically, recent advances \cite{hnnnips19} show that graph learning with Riemannian geometry presents superior expressiveness compared to the Euclidean counterpart and lower embedding distortions\footnote{Embedding distortion is measured by $\frac{1}{|\mathcal V|^2} \sum_{ij}\left| \frac{d_G(v_i, v_j)}{d(\mathbf x_i, \mathbf x_j)}-1\right|$, where each node $v_i \in \mathcal V$ is embedded as $\mathbf x_i$ in representation space. $d_G$ and $d$ denote the distance in the graph and the space, respectively.}.  
Incorporating Riemannian geometry has emerged as a promising direction for learning on graphs.

In this paper, we introduce the concept of Riemannian graph learning and discuss the difference and connection to some related topics.

\begin{definition}[{\textbf{Riemannian Graph Learning}}]
It refers to a novel family of neural networks that incorporate the Riemannian metric underlying graph structures and achieve the intrinsic geometric properties of Riemannian manifolds through differentiable neural operations or network layers. 
\end{definition}

\noindent We emphasize that, instead of introducing formulas in classic Riemannian geometry, the core of Riemannian graph learning lies in \textbf{endowing neural networks with the manifolds' intrinsic properties} such as conformal, equivariant, quotient, and symmetry, through novel differentiable formulations, which are difficult to achieve even with careful design in ordinary Euclidean geometry.
Although Euclidean geometry is indeed a special case of Riemannian geometry, Riemannian graph learning focuses on a wide family of non-Euclidean manifolds.
We use the term "Riemannian" to highlight the well-defined, intrinsic properties rooted in Riemannian manifolds.

We further distinguish Riemannian graph learning from related paradigms, such as geometric deep learning, metric learning, and hyperbolic deep learning. While these related fields often address broader domains, Riemannian graph learning specifically focuses on graph-structured data and the methodological nuances that bridge or differentiate these approaches.
\begin{figure}
    \centering
    \includegraphics[width=0.8\linewidth]{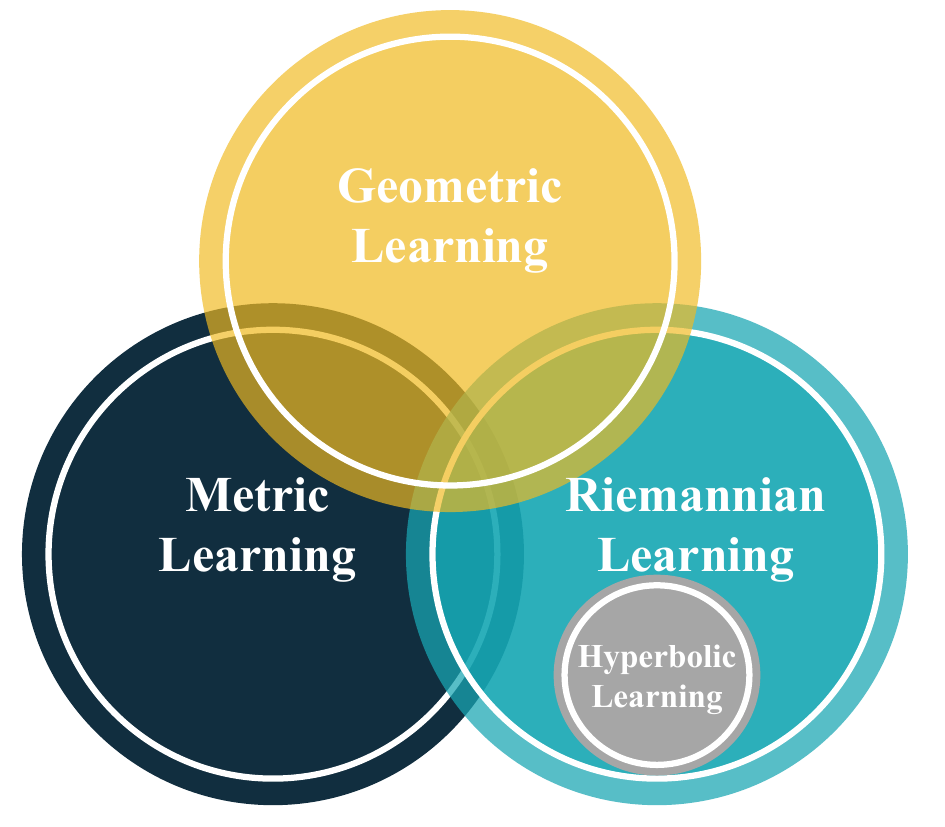}
    \caption{Relationship between Riemannian learning and some related ones.}
    \label{fig:relation}
    \vspace{-6mm}
\end{figure}

\begin{itemize}
    \item \textbf{Metric Learning.}

Instead of using Euclidean distance or Cosine similarity, metric learning is concerned with learning a distance function tuned to a specific learning task. 
Riemannian graph learning admits the geometric prior and works with Riemannian metrics.
Rather than a fixed distance, the distance function is learned in the framework of Riemannian geometry, offering geometric interpretation at the same time.  
That is, the essential difference lies in the \emph{admission of geometric priors}, and both approaches provide the flexibility to learn the similarity in a data-driven fashion.

 \item \textbf{Geometric Deep Learning.}
It primarily models the 3D coordinates, such as point clouds or meshes, and concerns about the geometric quantities such as distance and angles.
In terms of representation space, geometric deep learning typically works with the  Euclidean 3D space, while the focus of Riemannian graph learning is on non-Euclidean spaces.
In essence, geometric deep learning models the \emph{extrinsic} geometric conformation, but Riemannian graph learning explores the \emph{intrinsic} manifold properties.

 \item \textbf{Hyperbolic Deep Learning.}
It extends the traditional neural networks to the hyperbolic space.
Hyperbolic deep learning has gained growing attention in recent years owing to its off-the-shelf closed formulation, and  can be generally regarded as a subset of Riemannian deep learning. 
We emphasize that there exists a wide family of Riemannian manifolds beyond hyperbolic spaces, such as spherical spaces and matrix manifolds of Grassmann manifold and Symmetric Positive Definite manifold, and each of them comes with fruitful geometric advantages. 
\end{itemize}

\noindent The relationship among the aforementioned topics is sketched in Fig. \ref{fig:relation}.

\subsection{Related Surveys}
Numerous surveys have reviewed graph learning in Euclidean space.
Comprehensive overviews of Graph Neural Networks (GNNs) and their architectures (e.g., spectral and spatial GNNs) are provided in \cite{chen2020graph, cai2018comprehensive,ju2024comprehensive, khoshraftar2024survey,wu2020comprehensive}. Beyond general architectures, recent surveys have focused on specific tasks, such as knowledge graphs \cite{ji2021survey}, graph clustering\cite{liu2022survey}, and graph kernels\cite{kriege2020survey}.  Additionally, advanced learning paradigms, including self-supervised learning\cite{liu2022graph} and heterophilic graph learning \cite{zheng2022graph}, have been extensively summarized. 
However, these works predominantly operate under the assumption of a flat Euclidean latent space, largely overlooking the intrinsic geometry of complex data structures. 

Recognizing the limitations of Euclidean embeddings, recent surveys have begun to explore geometric deep learning. However, these studies primarily narrow their focus to hyperbolic geometry, treating it as the sole representative of non-Euclidean approaches \cite{mettes2024hyperbolic,fang2023hyperbolic,yang2022hyperbolic,peng2021hyperbolic}. While hyperbolic space is effective for hierarchical data, it represents only a fraction of the broader Riemannian landscape. 

Unlike existing works that are confined to hyperbolic space, this survey offers a holistic perspective on the entire Riemannian family. We systematically review diverse manifold types, including hyperbolic, spherical, product, and quotient, and others. This fills a critical gap by establishing a systematic reference for aligning diverse graph structural patterns with their optimal geometric spaces.

\section{Preliminaries} \label{sec: preliminaries}

Important notations are summarized in Table~\ref{notation}.
\begin{table}[t]
\caption{Notation Table.}
\centering
\resizebox{\columnwidth}{!}{
\begin{tabular}{|c|c|}
    \hline
    \textbf{Notation} & \textbf{Description} \\
    \hline
    $G = (V, E)$ &  Graph consisting of vertex set $V$ and edge set $E$.  \\
    $\mathbf{A} $ &  Graph adjacency matrix. \\
    $\boldsymbol x / \mathbf{x}$ & Vector or point on manifold. \\
    $\mathbf{X} $ &  Node attributes matrix. \\
    $\mathbf{h}_i$ & Embedding of node $i$. \\
    $\mathcal{L}$ & Loss function. \\
    $\mathbf{W}$  & A learnable weight matrix. \\
    $\alpha(\cdot,\cdot)$ & Attention score. \\
    $\mathcal{N}(\cdot, \cdot)$ & The Gaussian distribution.\\
    $\odot, \otimes$ & The Hadamard, Cartesian product.\\
    $\oplus$ & The Direct Sum. \\
    $\operatorname{Agg}(\cdot)$& A general Aggregation function.\\
    $\sigma(\cdot) $ & A nonlinear activation function.\\
    \hline
    
    $\mathcal{M}$ & Manifold $\mathcal{M}$. \\
    $\mathcal{G}{r}(k, n)$ & Grassmann manifold of $k$-planes in $\mathbb{R}^n$. \\
    $\mathcal{S}^n_{++}$ &  SPD manifold of $n \times n$ matrices.\\
    $\mathcal{T}_p\mathcal{M}$ & Tangent space of manifold $\mathcal{M}$ at point $p$. \\
    $\mathcal{T}\mathcal{M}$ & Tangent bundle of the manifold $\mathcal{M}$. \\
    $PT_{\cdot \rightarrow \cdot}$ & The Parallel Transport operator. \\
    $g$ & Riemannian metric. \\
    $\mathcal{G}$ & Lie group. \\
    $\mathfrak{g}$ & Lie algebra of Lie group. \\
    $ \gamma $ & A curve $\gamma$. \\
    $\kappa$ & Curvature. \\
    $ K $ & Gaussian curvature. \\
    $\operatorname{Ric}(\cdot,\cdot)$ & Ricci curvature value.\\

    $ \mathbb{E}^{n} / \mathbb{R}^d $ &\(n\)-dimensional euclidean space. \\
    $ \mathbb{H}^{n} $ &\(n\)-dimensional hyperbolic space. \\
    $ \mathbb{S}^{n} $ &\(n\)-dimensional spherical space. \\
    $\mathfrak{st}_{\kappa}^{d}$ & The $\kappa$-stereographic model space of dimension $d$.\\
    
    $\langle \cdot, \cdot \rangle$ & Euclidean inner product. \\
    $\langle \cdot, \cdot \rangle_\mathcal{L}$ & Lorentz inner product. \\
    $W_{1}(\cdot,\cdot)$ & Wasserstein distance. \\

    $\log^\kappa_{p}(\cdot) /\operatorname{Log}(\cdot)$ & The logarithmic map at point $p$ with curvature $\kappa$.\\
    $\exp^\kappa_{p}(\cdot) /\operatorname{Exp}(\cdot)$ & The exponential map at point $p$ with curvature $\kappa$.\\

    $\oplus_\kappa$ & The Möbius addition in the Gyrovector Space algebra. \\
    
    \hline
\end{tabular}
}
\label{notation}
\end{table}

\subsection{Riemannian Manifold and Geodesic}
A $d$-dimensional Riemannian manifold $(\mathcal{M},g)$ is a smooth manifold equipped with a Riemannian metric $g$. For any point $x \in \mathcal{M}$, the metric $g$ defines a positive-definite inner product $g_x(\cdot, \cdot): \mathcal{T}_x\mathcal{M} \times \mathcal{T}_x\mathcal{M} \to \mathbb{R}$ on the tangent space $\mathcal{T}_x\mathcal{M}$. This metric allows for the rigorous definition of local geometric quantities, such as angles, lengths, and volumes. 

Geodesics generalize the concept of straight lines to curved spaces. Formally, a geodesic is a curve that locally minimizes the distance between two points on the manifold. The intrinsic geometry of $\mathcal{M}$ determines the behavior of geodesics. In Euclidean space, geodesics are straight lines, and the sum of the interior angles of a triangle is exactly $\pi$. In curved spaces, however, this property no longer holds. For a geodesic triangle on a surface with Gaussian curvature $K$, the Gauss–Bonnet theorem \cite{demasongauss} gives $\alpha + \beta + \gamma - \pi = \int_{\triangle} K \, dA, $
where $\alpha, \beta, \gamma$ are the interior angles of the triangle, and $dA$ is the area element.
The deviation of the angle sum from $\pi$ thus equals the total curvature enclosed by the triangle, reflecting the intrinsic geometry of the space.

\subsection{Tangent Space and Tangent Bundle}
For any point $p \in \mathcal{M}$, the tangent space $\mathcal{T}_p\mathcal{M}$ is an $d$-dimensional vector space containing the tangent vectors of all curves passing through $p$. Intuitively, it serves as a local Euclidean approximation of the manifold. Since curved manifolds lack vector space properties (e.g., vector addition is undefined), most neural operations, such as aggregation and feature transformation are performed within this flat tangent space. 

The tangent bundle $\mathcal{T}\mathcal{M}$ unifies all local tangent spaces into a global structure, formally defined as:
\begin{equation}
    \mathcal{T} \mathcal{M} := \bigsqcup_{p \in M} \mathcal{T}_p\mathcal{M},
\end{equation}
The tangent bundle is a $2d$-dimensional smooth manifold where each point is a pair $(p,v)$, representing a position $p$ and a tangent vector $v \in \mathcal{T}_p\mathcal{M}$.

\subsection{Levi-Civita Connection and Curvature}
Since tangent spaces at different points are distinct vector spaces, vectors cannot be directly compared or aggregated across the manifold. The \textbf{Levi-Civita connection $\nabla$ }provides a canonical way to connect these spaces. And it defines the parallel transport, a mechanism that transports a vector from a point's tangent space to another point along a geodesic while preserving its norm and direction relative to the manifold's geometry.

Curvature quantifies how the manifold deviates from flat Euclidean space. 

\textbf{Sectional curvature} generalizes Gaussian curvature to higher-dimensional manifolds.
It measures the curvature of a two-dimensional plane inside the tangent space at a given point.
Aggregating sectional curvatures yields the \textbf{Ricci curvature}, which measures the average deformation of geodesic volume in different directions.
These curvature measures collectively describe the manifold’s local geometric behavior and provide insights into its global structure and topology.

\subsection{Lie Groups, Lorentz Group, and Equivariance}
A Lie group is a mathematical structure that is simultaneously a group and a smooth manifold. This duality allows it to describe continuous symmetries mathematically.  In geometric deep learning, Lie groups serve as the foundational tool for modeling transformations, such as rotations or translations, in a differentiable manner. 
The Lorentz group, which originating from special relativity, defines the the isometry group of the Hyperbolic space. Just as the orthogonal group defines rotations in Euclidean space, the Lorentz group describes rotations that preserve the hyperbolic structure (Minkowski metric). Its two basic elements are the Lorentz boost (a generalization of translation) and Lorentz rotation. The  Lorentz rotations form an SO(3) subgroup of the Lorentz group.
These elements allow for clean matrix representation and decomposition, which are essential for defining hyperbolic graph neural network layers. 

Equivariance is a key concept when dealing with Lie groups, which ensures that the structure of the data is respected during processing. Formally, equivariance means that the function commutes with the transformation: applying the transformation before the function yields the same result as applying it after (i.e., $f(T(x))=T(f(x))$).

\subsection{Lie Algebra, Exponential and Logarithm Maps}

Every Lie group \( \mathcal{G} \) has an associated Lie algebra \( \mathfrak{g} \), which is the tangent space at the identity element. 
Since \( \mathfrak{g} \) is a vector space, it serves as a linear approximation of the group, enabling efficient computations. The transition between the group and its algebra is achieved  by the Lie exponential map $\exp: \mathfrak{g} \rightarrow \mathcal{G}$ and its inverse, the logarithm map $\log: \mathcal{G} \rightarrow \mathfrak{g} $.

In the broader context of Riemannian geometry, these maps serve as essential bridge between the manifold $\mathcal{M}$ and the flat tangent space $\mathcal{T}_p \mathcal{M}$. 
Given a point \( p \in \mathcal{M} \) and tangent vector \( v \in \mathcal{T}_p \mathcal{M} \), the logarithm map projects $p$ into the tangent space at $p$, represented as the tangent vector $v$. 
Conversely, the exponential map projects a tangent vector back onto the manifold along a geodesic path.

\subsection{Diffeomorphism}
In Riemannian geometry, a diffeomorphism is a smooth, bijective map $f: \mathcal{M} \to \mathcal{N}$ between two smooth manifolds such that its inverse $f^{-1}$ is also smooth.
The existence of such a map implies $\mathcal{M}$ and $\mathcal{N}$ are isomorphic as smooth manifolds, i.e., they are indistinguishable from the perspective of differential geometry.  

\subsection{Cohomology}
Cohomology is a central tool in algebraic topology that characterizes the global structure of a manifold by identifying and classifying its ``holes''. These holes represent features that cannot be continuously deformed (contracted) to a point. Cohomology assigns algebraic invariants, specifically cohomology groups, to quantify the number and dimension of these topological features. In graph learning, this tool provides a way to move beyond simple connectivity and to analyze and leverage high-order organization, such as cycles or cavities, within the graph structure. 
\subsection{The Cartan Moving Frame Method}
The Cartan Moving Frame Method is a powerful computational framework in differential geometry. Instead of relying on a fixed global coordinate system, it attaches a tailored local basis to each point on the manifold and uses exterior differential forms to describe how these frames evolve and rotate across the space. In the context of Riemannian geometry and graph learning, its primary advantage is being coordinate-free. This approach significantly simplifies the computation of geometric invariants, such as curvature and connection, thereby facilitating a more intuitive and efficient analysis of the intrinsic geometric structures of manifolds.

\subsection{Chern Classes}
Chern classes are fundamental algebraic invariants in differential geometry, specifically designed to characterize the global topological properties of complex vector bundles. Constructed from local curvature forms via integration, they transform microscopic geometric bending into discrete values that reflect the global structure. Geometrically, Chern classes quantify the non-triviality of the bundle. Specifically, a non-zero Chern class indicates a topological obstruction, signifying that the vector bundle cannot be globally trivialized. 

\section{Taxonomies} \label{sec: Taxonomies}
This section provides a clear roadmap for Riemannian machine learning. It covers the basic theories behind Riemannian manifolds, the design of Riemannian neural network structures, and specialized learning methods for non-Euclidean data. It focuses on how the geometric properties of manifolds drive architectural innovations and specialized learning strategies. The research taxonomy is structured into three sections, outlined below.
An overview of the proposed taxonomy is illustrated in Fig. \ref{pttaxonomy}, and a continuously updated classification with related resources is available online.\footnote{\url{https://github.com/wq177/Awesome_Riemannian_Graph_Learning}}

\subsection{Manifolds Type}

The family of Riemannian manifolds is a class of smooth manifolds equipped with a Riemannian metric. It models the geometric properties of data through intrinsic geometric structures and is widely used in machine learning and graph learning. Based on differences in geometry and structures, we categorize Manifolds into eight aspects:

\begin{enumerate}
\item \textit{Hyperbolic Manifold}: It has constant negative curvature (\(\kappa < 0\)). Its distance expands exponentially, making it particularly suitable for representing hierarchical or tree-like data (e.g., graph-structured data).

\item \textit{Spherical Manifold}: It has constant positive curvature (\(\kappa > 0\)). It naturally captures cyclical patterns and is an essential for tasks requiring directional or rotational symmetry.

\item \textit{Constant Curvature Space (CCS)}: Instead of manually fixing the geometry, CCS serves as a unified framework where the curvature $\kappa$ is a learnable parameter.

\item \textit{Product and Quotient Space}: A Product space is formed by the Cartesian product of simpler manifolds (e.g., $\mathbb{H}^n \times \mathbb{S}^m$). They are tailored for graph data containing mixed topologies.
A Quotient Space is formed by dividing the original space into equivalence classes via equivalence relations.

\item \textit{Pseudo-Riemannian Manifold}: Unlike standard Riemannian manifolds, these spaces utilize an indefinite metric, accommodating both space-like and time-like dimensions.  It is suitable for modeling directed graphs.

\item \textit{Grassmann Manifold}: It consists of all k-dimensional linear subspaces in n-dimensional space. By modeling data collections as points on the manifold, it can compactly capture the intrinsic structure of high-dimensional data collections.

\item \textit{SPD Manifold}: It is composed of all \(n \times n\) symmetric positive definite matrices. Its modeling relies on Riemannian metrics. It is used in machine learning tasks like dimensionality reduction and classification.

\item \textit{Generic Manifold}: Rather than embedding into a predefined model space, this category treats the graph itself as a discretized manifold. Ricci curvatures have been used in graph data modeling, improving the performance of traditional methods in clustering and information diffusion.
\end{enumerate}

The family of Riemannian manifolds provides accurate geometric modeling tools for data with different properties. In current research, various manifolds have been widely applied in deep learning: Hyperbolic manifolds have promoted the development of hierarchical data embedding models; spherical manifolds have optimized feature constraints and symmetry modeling. However, the application of some manifolds (e.g., SPD manifolds) still needs further exploration. Future work should focus on cross-manifold fusion and adaptation to complex scenarios.

\subsection{Riemannian Neural Architectures} Riemannian Neural Architectures extend traditional neural networks to Riemannian manifolds. They leverage intrinsic geometric properties of manifolds to model data with non-Euclidean structures. These architectures are mainly categorized into seven types.
\begin{enumerate}

\item \textit{Riemannian Convolutional Network}: Built on the message-passing of graph convolutional networks, its key is to aggregate neighbor features on manifolds. It processes manifold-valued features through tangent space aggregation or direct manifold aggregation.

\item \textit{Riemannian Variational Autoencoder}: It addresses the challenge of modeling latent distributions in non-Euclidean spaces. It employs Riemannian normal distributions and wrapped normal distributions as latent distributions, supporting learning in mixed-curvature spaces.

\item \textit{Riemannian Transformer}: Extends the Transformer architecture to Riemannian manifolds, with the core being adapting self-attention to manifold curvature and topology. 

\item \textit{Riemannian Graph ODE}: Combines ordinary differential equations (ODEs) with Riemannian geometry to model continuous dynamic evolution of data on manifolds. It has been applied to physical system modeling and information diffusion prediction \cite{Pioneer2025AAAI}.

\item \textit{Riemannian Denoising Diffusion and SDE}: Extending stochastic differential equations (SDEs) to Riemannian manifolds, it introduces manifold Brownian motion to model stochastic dynamical systems. It enables uncertainty-aware representation learning for graph-structured data.

\item \textit{Riemannian Flow Matching}: Extending invertible flow models to Riemannian manifolds, it uses continuous normalizing flows to model complex distributions. It addresses the theoretical flaws of early flow models that relied on Euclidean projections.
\end{enumerate}
The goal of Riemannian Neural Architectures is to overcome the limitations of Euclidean spaces. Current research has made significant progress: mainstream architectures have been extended to Riemannian settings; and applications in graph learning, generative modeling, and physical system prediction have been shown.

\subsection{Learning Paradigms}

It outlines learning paradigms customized for Riemannian manifolds. They are mainly divided into four categories:

\begin{enumerate}
\item \textit{Unsupervised Learning}: It does not require labeled data and focuses on mining the inherent structure of graphs for clustering or representation learning. 

\item \textit{Semi-supervised Learning}: Combines limited labeled data with abundant unlabeled structures. Riemannian geometry enhances this paradigm by using the manifold geometry as a strong inductive bias.

\item \textit{Self-supervised Learning}: It learns effective graph representations without manual labels. The reconstructive type learns by reconstructing missing graph links; the contrastive type uses contrastive objectives to learn robust features.

\item \textit{Foundation Model and Transfer Learning}: Graph Foundation Models (GFMs) achieve cross-domain knowledge transfer through large-scale pre-training. Pre-trained GFMs can be fine-tuned or adapted to specific downstream tasks across various graph domains.
\end{enumerate}
Graph learning paradigms have enhanced graph modeling capabilities through diverse strategies, with foundation models and transfer learning being enhanced research hotspots. Add reference is the first to introduce Riemannian geometry, opening up new directions for complex graph modeling.

\section{Family of Riemannian Manifolds} \label{sec: manifold}
\begin{table*}[htbp]
\centering
\caption{Summary of Manifold. }
\begin{tabular}{l c p{10.0cm}}
\toprule
\textbf{Manifold} & \textbf{Model Space} & \textbf{Method} \\
\midrule
\multirow{2}{*}{Hyperbolic} 
  & Poincaré
    & HGNN\cite{hnnnips19}, 
    p-VAE\cite{mathieu2019continuous},
    ROTE\cite{ATTH20},
    HTF\cite{HTF},
    HGCF\cite{sun2021hgcf},
    DT-GCN\cite{DT-GCN21},
    ACE-HGNN\cite{ace-hgnn2021},
    CurvGAN\cite{li2022curvature},
    HIE\cite{hie23icml},
    HDD\cite{LinCMT23},
    HGWaveNet\cite{bai2023hgwavenet},
    HyperIMBA\cite{HyperIMBA23},
    SHAN\cite{li2023multi}, 
    FFHR\cite{FFHR},
    H-Diffu\cite{FengZFFWLS23},
    HypMix\cite{HypMix24},
    HMPTGN\cite{LeT24},
    MSGAT\cite{msgat2024icdm}, 
    HGCH\cite{HGCH24},
    THGNets\cite{liu2025thgnets}, 
    HGDM\cite{HGDM24AAAI},
    HyperDefender\cite{MalikGK25},
    HyperGCL\cite{HyperGCL24kdd},
    HEDML\cite{HEDML25www}
         \\
  & Lorentz
  & HGCN\cite{hgcn2019nips},
  HGNN\cite{hnnnips19}, 
  $\mathcal{TC}$-flow \cite{BoseSLPH20}, 
  WHC\cite{BoseSLPH20},
  HVGNN\cite{tgnn2021}, 
  LGCN\cite{LGCN21WWW},
  H2H-GCN\cite{h2hgcn2021},
  RotDif\cite{QiaoFLLH0Y23},
  R-HGCN\cite{RHGCNXue2024AAAI}, 
  Hypformer\cite{yang2024hypformer}, 
  LorentzKG\cite{LorentzKG}, 
  LSEnet\cite{sunLSEnet}, 
  HypDiff\cite{HypDiff24},
  MHR\cite{feng2025mhr},
  Hgformer\cite{LHGCN},
  HVGAE\cite{liu2025hyperbolic}
    \\
\midrule
\multirow{1}{*}{Constant Curvature} 
    & -- 
    &DyERNIE\cite{DyERNIE},
    MVAE\cite{skopek2019mixed},
    M2GNN\cite{wang2021mixed},
    SelfMGNN\cite{sun2022aaai},
    HGE\cite{pan2024hge},
    IME\cite{wang2024ime},
    MCKGC\cite{gao2025mixed}
    \\
\midrule    
\multirow{1}{*}{Spherical} 
  & -- 
    & DeepSphere\cite{defferrard2020deepsphere}, 
    LHML \cite{guo2022learning}, 
    SMP\cite{smp22iclr},
    Q-align\cite{zhu2022spherical}\\
\midrule
\multirow{1}{*}{Product and Quotient Space} 
  & -- 
    & $\kappa$-GCN\cite{Bachmann2020ICML}, 
    DyERNIE\cite{DyERNIE},
    MAVE\cite{skopek2019mixed},
    M2GNN\cite{wang2021mixed},
    SelfMGNN\cite{sun2022aaai},
    HGE\cite{pan2024hge}, 
    ProGDM\cite{ProGDM24},  
    IME\cite{wang2024ime},
    GeoMancer\cite{Geomancer25gao},
    GraphMoRE\cite{GraphMoRE},
    MCKGC\cite{gao2025mixed},
    RiemannGFM\cite{RGFM25sun} \\
\midrule
\multirow{1}{*}{Pseudo-Riemannian Manifold} 
  & --
    & 
    PseudoNet\cite{law2020ultrahyperbolic},
    UltraNet\cite{law2021ultrahyperbolic},
    Directed-Pseudo\cite{sim2021directed}, 
    $\mathcal{Q}$-GCN\cite{Xiong2022NeurIPS}, UltraE\cite{xiong2022ultrahyperbolic},  
    $\mathcal{Q}$-GT\cite{qgt25}\\
\midrule
\multirow{1}{*}{Grassmann Manifold} 
  & -- 
    & EGG\cite{EGG24zhou},
    GyroGr\cite{GyroGr},
    SPSD\cite{GyroSpsd++24son},
    CAGM\cite{ma2024cross},
    sLGm\cite{wu2024multiple},
    CLRSR\cite{wang2018cascaded}, G-ALDNLR\cite{piao2019double}, PGM-HLE\cite{batalo2022analysis}, GrCNF\cite{yataka2023grassmann}, GRLGQ\cite{mohammadi2024generalized}, AMCGM\cite{wu2022attention} \\

\midrule
\multirow{1}{*}{SPD Manifold} 
  &--
    & Graph-CSPNet\cite{ju2024graph}, 
    GyroGr\cite{nguyen2023building},
    SPSD\cite{GyroSpsd++24son},
    GBWBN\cite{wang2025learning},
    SPDNet\cite{huang2016riemannian}, DMTNet\cite{zhang2017deep}, SPDNetNAS\cite{sukthanker2021neural}, SymNet\cite{wang2022symnet}, MSNet\cite{chen2023riemannian} \\
\midrule
\multirow{1}{*}{Generic Manifold} 
  & --
    &RicciNet\cite{sunRiccinet}, \textsc{Congregate}\cite{sunCONGREGATE}, R-ODE\cite{sun2024r}, MSG\cite{MSG2024nips}, MofitRGC\cite{sun2024aaai}, Pioneer\cite{Pioneer2025AAAI},CurvGAD\cite{CurvGAD}, GBN \cite{gbn25sun}  \\
 
\bottomrule
\end{tabular}
\end{table*}

\subsection{Hyperbolic Space}

\begin{figure}
    \centering
    \includegraphics[width=\linewidth]{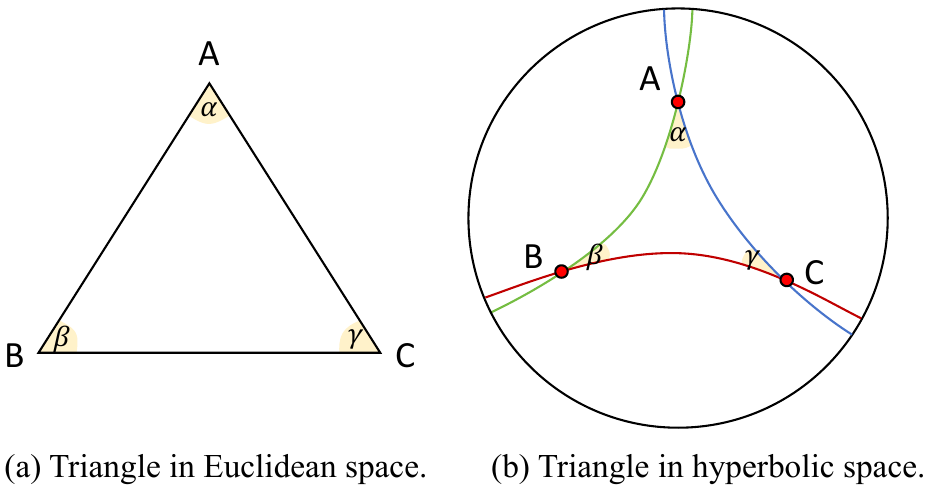}
    \caption{Comparison between Euclidean (flat) and Hyperbolic (negatively curved) triangles.}
    \label{fig:triangle}
    \vspace{-7mm}
\end{figure}

As a negatively curved Riemannian manifold, hyperbolic space gains fundamental importance through its application in Einstein’s
special theory of relativity.
We showcase a geometric intuition of negative curving. 
As shown in Fig. \ref{fig:triangle}, hyperbolic space bends geodesic triangles curved inward versus the straight edges in the flat geometry, i.e., Euclidean space.
Hyperbolic space can be embedded as a subspace of high-dimensional $\mathbb R^d$ space, which is known as \emph{model space}.
The popular model spaces include the Poincar\'e model, Lorentz model, Klein model and Poincar\'e half-plane model. 

The geometric properties of hyperbolic space are deeply linked to statistical manifolds. For instance, the parameter space of univariate Gaussian distributions equipped with the Fisher Information Metric (FIM) is isometric to the Poincar\'e half-plane. In this setting, the \emph{Fisher-Rao distance} $d_{FR}$ defines the geodesic distance between distributions, which is locally coupled with the Kullback-Leibler (KL) divergence as $D_{KL}(P\|Q) \approx \frac{1}{2} d_{FR}^2(P, Q)$. This connection allows KL divergence to serve as a computationally efficient proxy for hyperbolic distances in uncertainty-aware representation learning.

\textbf{Lorentz Model (a.k.a. Hyperboloid Model)}. 
The $d$-dimensional Lorentz model $\mathbb{H}^{d}$ is defined on the $(d+1)$-dimensional manifold satisfying $\left \langle x,x \right \rangle_{\mathcal{L}}=1/\kappa$ with $x_0>0$. It utilizes the Minkowski inner product $\left \langle \mathbf{x},\mathbf{y} \right \rangle_{\mathcal{L}}=-x_0y_0 + \sum_{i=1}^{d} x_i y_i$. The distance is given by $d_\mathbb{H}=\frac{1}{\sqrt{|\kappa|}} \operatorname{arccosh}\left(\kappa\langle x, y\rangle_{\mathcal{L}}\right)$

$\textbf{Poincar\'e Model.}$ 
The Poincar\'e model $\mathbb{B}_{\kappa}^d$ is defined within the open ball $\left \{ x \in \mathbb{R}^d|  \left \|  x\right \|<1/\sqrt{|\kappa|} \right \} $. Its metric is conformal to the Euclidean metric $g_x^{\mathbb{B}} = \lambda_x^{2}g_x^E$ with the conformal factor $\lambda_{x}=\frac{2}{1+\kappa\|x\|^{2}}$. The distance is calculated as $d_{\mathbb{B}}^{\kappa}=\frac{1}{\sqrt{|\kappa|}} \cosh ^{-1}\left(1+\frac{2 \kappa\|\boldsymbol{x}-\boldsymbol{y}\|^{2}}{\left(1+\kappa\|\boldsymbol{x}\|^{2}\right)\left(1+\kappa\|\boldsymbol{y}\|^{2}\right)}\right)$.


$\textbf{Klein Model.}$ 
The Klein model $\mathbb {K}^n_\kappa$ shares the same domain as the Poincar\'e ball but employs a different metric structure. The Riemannian metric is $g_x(u,v)=\frac{\langle v,v\rangle}{1-\|x\|^2}+\frac{\langle x,v\rangle^2}{(1-\|x\|^2)^2}$. Its distance is derived as  $d_{\mathbb{K}}(x, y) = \frac{1}{\sqrt{|\kappa|}}\operatorname{arcosh}\left( \frac{1 - \kappa\langle x, y \rangle}{\sqrt{(1 - \kappa\|x\|^2)(1 - \kappa\|y\|^2)}} \right)$.

$\textbf{Poincar\'e half-plane.}$ The Poincar\'e half-plane is defined as $\mathbb{U}^d = \left \{  \mathbf{x} \in \mathbb{R}^d | x_d>0\right \} $ with the Riemannian metric tensor $ds^2 = \frac{d\mathbf{x}^2_1 + d\mathbf{x}^2_2+ \cdots + d\mathbf{x}^2_n}{\mathbf{x}^2_n}$. The distance between two points $x,y \in \mathbb{U}^d$ is $d(x,y)=\operatorname{arcosh}\left(1 + \frac{\|x-y\|^2}{2 x_n y_n}\right)$.

Note that the model spaces of hyperbolic space are equivalent in essence and thus can be utilized interchangeably. However, each model space has its own numerical characteristics, e.g., the Lorentz model is numerically stable as reported in \cite{hgcn2019nips, LGCN21WWW}.
Next, we elaborate on the strengths of hyperbolic space.

\emph{(1) The hyperbolic space is associated with a powerful algebra tool -- the Lorentz group.} 
As a representative Lie group, the Lorentz group offers an elegant mathematical framework to study the transformations over hyperbolic space.
Specifically, the Lorentz group is a set of Lorentz transformations that preserve the Minkowski metric tensor.
In other words, Lorentz transformations are linear isometries where the Minkowski inner product is invariant under the transformation.
Consequently, it allows one to operate in hyperbolic space as naturally as in Euclidean space.
Lorentz boost\footnote{Lorentz boost is a linear transformation of the plane, and can be understood as a generalized translation under the special theory of relativity.} serves as a generalization of translation, while Lorentz rotation parallels its Euclidean counterpart.
They are the two basic elements of the Lorentz group, and each has a clean matrix representation subject to orthogonal constraints.
Also, any Lorentz transformation can be decomposed into a combination of Lorentz boosts and rotations.

\emph{(2) The hyperbolic space, which can be regarded as a continuous tree, has proven its superiority in representing hierarchical data.}
Intuitively,  hyperbolic space expands exponentially with respect to its radius, mirroring the growth of a tree \cite{hie23icml}. 
Theoretically, a tree can be embedded in low-dimensional hyperbolic space with bounded distortion, but this property does not hold in any Euclidean space \cite{sala2018representation}.
This demonstrates that, to some extent, trees and hyperbolic space exhibit a kind of isometry. 
The tree-likeness of hyperbolic space is frequently revisited in the literature.

In the literature, hyperbolic space has emerged as an impressive alternative to Euclidean space for defining graph convolution networks. 
Recent studies have explored various architectures that integrate hyperbolic geometry into graph learning.
HGNN \cite{hnnnips19} introduces a general graph neural network framework on Riemannian manifolds.
HGCN \cite{hgcn2019nips} extends traditional GCNs to hyperbolic space, enabling the effective representation of hierarchical graphs.
HGAT \cite{HGAT2021} exploits attention mechanism for hyperbolic node information propagation and aggregation.
HGDM \cite{HGDM24AAAI} exploits hyperbolic latent spaces for diffusion modeling of hierarchical or power-law graphs.
P-VAE \cite{mathieu2019continuous} adapts variational auto-encoders to a Poincaré ball latent space in order to better capture tree-like or hierarchical data distributions.
Hyperbolic-PDE GNNs \cite{yue2025hyperbolic-pdeGNN} reinterpret spectral graph neural networks as systems of hyperbolic partial differential equations, bridging geometry and spectral analysis.
LSEnet \cite{sunLSEnet} employs the Lorentz model and structural entropy to achieve unsupervised graph clustering in hyperbolic space.
It demonstrates the practical utility of hyperbolic space through applications in various domains.
So far, most existing works primarily consider the hyperbolic representation learning, while there exist opportunities for exploring the implicit tree structure in hyperbolic space.


\subsection{Hyperspherical Space}

\begin{figure}
    \centering
    \includegraphics[width=\linewidth]{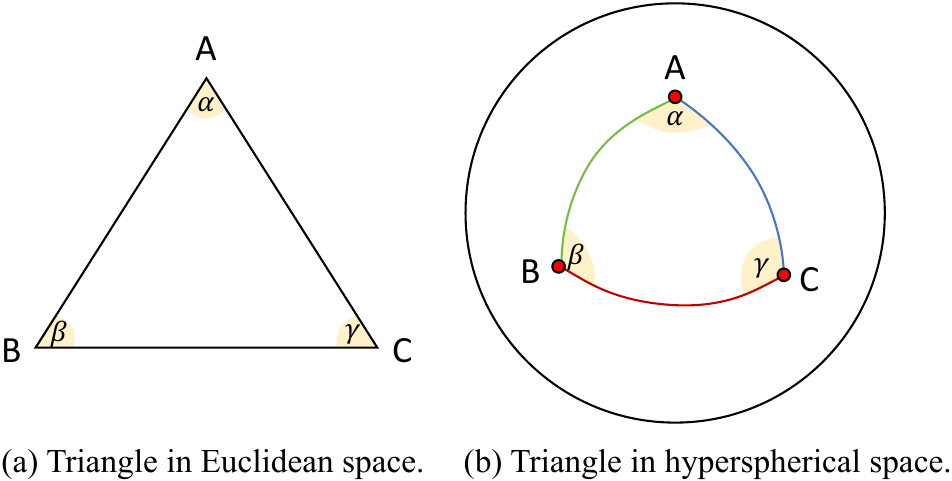}
    \caption{Comparison between Euclidean (flat) and Hyperspherical (positively curved) triangles.}
    \label{fig:triangle2}
    \vspace{-7mm}
\end{figure}

As a positively curved Riemannian manifold, hyperspherical space\footnote{It is also known as an elliptic manifold, while we utilize the term hyperspherical space throughout this survey following the convention in machine learning community.} is the high-dimensional generalization of $3$-dimensional sphere, and it renders geodesic
triangles curved outward as illustrated in Fig. \ref{fig:triangle2}.

$\textbf{Sphere Model.} $ The sphere model $\mathbb{S}^d_{\kappa}=\left \{ x \in \mathbb{R}^d|\left \|x  \right \|^2=1/\kappa  \right \} $ is defined as a submanifold embedded in the ($d$+1)-dimensional manifold. The metric is strictly induced from the ambient Euclidean inner product  $\left \langle \mathbf{x}, \mathbf{y}\right \rangle = \sum_{i=0}^{d} x_i y_i $. The distance between two points $x,y$ is $d_{\mathbb{S}}(x,y) = \frac{1}{\sqrt{\kappa}}\operatorname{arccos} (\kappa\left \langle \mathbf{x}, \mathbf{y} \right \rangle )$. 

The utility of hyperspherical space in graph learning is grounded in two intrinsic properties:

(1) Hyperspherical space is intrinsically linked to the special orthogonal group $SO(n)$, allowing rotational symmetry and directional information to be represented intrinsically rather than being extrinsically imposed. This connection to $SO(n)$ provides a mathematical foundation for building rotation-equivariant models for physical systems, such as molecular systems. 
(2) Hyperspherical space maintains strong discriminative ability for extremely high-dimensional data, which tends to be indistinguishable under Euclidean distance. 
This advantage is rooted in the superior stability of angular distance (or geodesic distance) in hyperspherical space \cite{textembedding/nips19}. 

In recent years, Spherical CNNs\cite{cohen2018spherical} introduced a network that is both expressive and rotation-equivariant. $\mathcal{S}$-VAE\cite{davidson2018hyperspherical} replaced the Gaussian distribution with the vMF distribution to adjust for data with hyperspherical structures. DeepSphere\cite{defferrard2020deepsphere}  modeled the sampled sphere as a graph, using graph convolutions to balance computational efficiency and rotation equivariance. 
Recently, several studies have leveraged the symmetries and other intrinsic properties of hyperspheres to achieve more effective representations and improve performance in various applications, such as molecular generation\cite{smp22iclr}, recommender systems\cite{zhu2022spherical}, and anomaly detection\cite{guo2022learning}. 
Most current studies introduce the hypersphere primarily to enhance modeling capacity, while overlooking how its inherent high symmetry can provide advantages such as translation and rotation equivariance, which hold great potential in fields such as molecular generation. 
\subsection{Constant Curvature Space}

While hyperbolic and spherical spaces provide distinct geometric inductive biases for hierarchical and cyclical data respectively, real-world graphs often exhibit heterogeneous topologies that cannot be adequately modeled by a fixed curvature.  To address this, the Constant Curvature Space (CCS) framework introduces a unified formalism, allowing models to seamlessly transition between Euclidean, hyperbolic, and spherical geometries by treating the curvature $\kappa$ as a learnable parameter. 

To unify these distinct geometries, CCS typically employs the  $\kappa$-stereographic model. For a curvature $\kappa$ and a dimension $d \ge 2$, the $\kappa$-stereographic model $\mathfrak{st}_{\kappa}^{d}$ is defined on the manifold of $\left \{ x \in \mathbb{R}^d|-\kappa \left \| x \right \|^2 <1  \right \} $, which is equipped with a Riemannian metric $\mathfrak{g}_{x}^{\kappa} = \frac{4}{(1+\kappa \left \|x  \right \|^2 )^2}\mathbf{I}$ for any curvature $\kappa$.  
It is defined as $\pi : \mathbb{H}_{\kappa}^{d}/\mathbb{S}_{\kappa}^{d} \to \mathfrak{st}_{\kappa}^{d}$ taking the form of:
\begin{equation}
    \pi: \mathbb{H}_{\kappa}^{d}/\mathbb{S}_{\kappa}^{d} \to \mathfrak{st}_{\kappa}^{d}, \quad \mathbf{x} = \frac{1}{1+\sqrt{|\kappa|}\mathbf{x}'_{d+1}}\mathbf{x}'_{1:d},
\end{equation}

For positive curvature, it can choose the stereographic projection of the sphere, while for negative curvature, it can choose the Poincar\'e model, which is the stereographic projection of the Lorentz model.

To perform neural operations (e.g., feature aggregation) in the space, CCS utilizes the Gyrovector Space algebra. Standard vector addition is replaced by M\"obius addition:
\begin{equation}
    \mathbf{x} \oplus_\kappa \mathbf{y} = \frac{(1 - 2\kappa \mathbf{x}^T \mathbf{y} - \kappa \|\mathbf{y}\|^2)\mathbf{x} + (1 + \kappa \|\mathbf{x}\|^2)\mathbf{y}}{1 - 2\kappa \mathbf{x}^T \mathbf{y} + \kappa^2 \|\mathbf{x}\|^2 \|\mathbf{y}\|^2},
\end{equation}
It is a generalized operator that respects the manifold's boundaries and curvature constraints, ensuring that node updates remain geometrically valid .

Recent years, $\kappa$-GCN\cite{Bachmann2020ICML}  extended the graph convolutional network to constant curvature spaces. 
CurvGAN \cite{li2022curvature} leveraged the unified framework to stabilize the training of GAN on graphs, dynamically adjusting curvature during the generation process. 
More recently, Pioneer\cite{Pioneer2025AAAI} utilized the $\kappa$-stereographic model to define continuous graph dynamics via Riemannian ODEs, while GraphMoRE\cite{GraphMoRE} employed a mixture of experts, assigning nodes to different local curvatures to handle extreme topological heterogeneity. Additionally, methods like CurvGAD\cite{CurvGAD} exploit curvature estimates for enhanced graph anomaly detection.

\subsection{Product and Quotient Manifolds}

Real-world graphs rarely exhibit a single, uniform geometry. To capture complex structures, Riemannian graph learning leverages composite manifolds constructed via product or quotient operations. 
\subsubsection{Product and Warped Product}
\begin{itemize}
    \item \textbf{Cartesian Product}:  Given a set of smooth manifolds $\mathcal{M}_1,\mathcal{M}_2, \cdots , \mathcal{M}_k$, the product manifold $\mathbb{P}$ is given as the Cartesian product of these manifolds:
\begin{equation}
    \mathbb{P} = \mathcal{M}_1 \times \mathcal{M}_2 \times \dots \times \mathcal{M}_k,
\end{equation}
where $\times$ denotes the Cartesian product. Specifically, with the Cartesian product construction, a point $\mathbf{x} \in \mathbb{P}$ are represented by a concatenation of $\mathbf{x} = [\mathbf{x}_1, \dots , \mathbf{x}_k]$. The tangent vector $\mathbf{v} \in \mathcal{T}_x\mathbb{P}$ at the point $\mathbf{x}$ is $\mathbf{v}=[\mathbf{v}_1,\dots, \mathbf{v}_k]$, where $\mathbf{v}_i \in \mathcal{T}_{x_i}\mathcal{M}_i$. For $\mathbf{x}$ and $\mathbf{y} \in \mathbb{P}$, the distance between them is defined as $d_{\mathbb{P}}(\mathbf{x},\mathbf{y}) = \sum_{i=1}^{k}d_{\mathcal{M}_i}(\mathbf{x}_i,\mathbf{y}_i)$. 
\item \textbf{Warp Product:} Given two Riemannian manifolds $(\mathcal{B}, g_{\mathcal{B}})$ and $(\mathcal{F}, g_{\mathcal{F}})$, and a positive function $w:\mathcal{B}\rightarrow\mathbb{R}_{>0}$, the warped product $\mathcal{M}=\mathcal{B}\times_{w}\mathcal{F}$ is equipped with the metric
\begin{equation}
    g = g_{\mathcal{B}} \oplus w^{2}(b)\, g_{\mathcal{F}}, \quad b\in\mathcal{B}.
\end{equation}

This formulation creates manifolds with heterogeneous curvature profiles that vary across the space, offering richer expressiveness than standard products. 

\end{itemize}

Recently, CUSP\cite{cusp} leverages a product manifold framework to integrate spectral graph theory with mixed-curvature geometry, independently learning and attentively fusing node representations across various curvature-curvature manifolds to capture complex, heterogeneous graph topologies. MixRPM\cite{gu2018learning} first introduced product manifolds to learn embedding and curvatures, and to estimate the product signature. MixVAE\cite{skopek2019mixed} extended the VAEs with product manifolds, enabling more effective modeling of complex data. In addition, product manifolds have shown great potential in other applications, including knowledge graphs\cite{pan2024hge,han2020dyernie,wang2021mixed,gao2025mixed,wang2024ime}, graph generation\cite{ProGDM24}, and graph foundation models\cite{RGFM25sun}. However, many existing approaches achieve model diversity by simply concatenating multiple models diversity, without deeply modeling the relationships between different component spaces.

\subsubsection{Quotient}

Quotient spaces address the issue of redundancy and symmetry.  
Given a smooth manifold $\mathcal{M}$ and a symmetry group $G$, the quotient manifold $\mathbb{Q}=\mathcal{M}/G$ treats all points $x \in \mathcal{M}$ that lie in the same orbit under the group action as a single point.

If the original manifold has a $G$-invariant Riemannian metric, a metric can be induced on the quotient space, ensuring well-defined geodesics and curvature. UltraNet\cite{law2021ultrahyperbolic} leverages equivalence classes in the quotient space to address optimization challenges caused by disconnected geodesics in ultra-hyperbolic space.

Although quotient manifolds have seen limited application in graph learning, their intrinsic advantages, such as eliminating redundant degrees of freedom and capturing symmetric structures, offer potential benefits for reducing model complexity and enabling equivariant molecular generation.

\subsection{Pseudo-Riemannian Manifold}


Standard Riemannian manifolds are constrained by positive-definite metrics, where the distance between distinct points is strictly positive, locally resembling Euclidean space. 
Pseudo-Riemannian manifolds relax this constraint by equipping the tangent space with a nondegenerate, indefinite metric tensor. 
Nondegeneracy means that $\forall \mathbf{u},\mathbf{v} \in T_{\mathbf{x}}\mathcal{M}, g_{\mathbf{x}}(\mathbf{u},\mathbf{v})=0$, then $\mathbf{u}=0$.  Indefinity means that the metric tensor could be of arbitrary signs. 

This generalization serves as the mathematical foundation for Einstein’s General Relativity, where the manifold is interpreted as spacetime—a connected, time-oriented Lorentz manifold. In this view, the manifold unifies spatial dimensions with a temporal dimension, providing a rigorous geometric framework for modeling events and their causal relationships. 

The defining characteristic of Pseudo-Riemannian manifolds is the indefinite metric, which classifies tangent vectors into four distinct causal characters: timelike (or chronological), null, spacelike or non-spacelike (or causal).
Unlike Riemannian geometry, where any two points are connected by a distance-minimizing geodesic, Pseudo-Riemannian geometry introduces a causal structure. This structure determines whether an event (a point on the manifold) can influence another. Specifically, a time-oriented manifold allows for the definition of a chronological order: an event $x$ precedes $y$ only if there exists a future-directed timelike curve connecting them. This intrinsic geometric property creates a distinction between reachable and unreachable regions, a feature absent in standard Riemannian manifolds.

This unique causal structure offers profound advantages for graph learning, particularly in modeling directed graphs. Pseudo-Riemannian spacetimes naturally encode directionality through chronological ordering: a directed edge $u \to v$ exists if and only if the embedding of $v$ lies in the future timecone of $u$. 

Recently, Directed-PNet\cite{sim2021directed} utilized the causal structure of Lorentzian spacetimes to embed directed graphs, interpreting edge directionality through lightcone constraints. Similarly,  $\mathcal{Q}$-GCN\cite{Xiong2022NeurIPS} and UltraE \cite{xiong2022ultrahyperbolic} extended Graph Convolutional Networks (GCNs) and Knowledge Graph embeddings to Ultrahyperbolic manifolds, demonstrating their capability to model heterogeneous topologies that contain both hierarchies and cycles. 

\subsection{Grassmann Manifold and Stiefel manifold}
While most Riemannian approaches map entities to point vectors, the Grassmann Manifold $\mathcal{G}r(k,n)$ models entities as $k$-dimensional linear subspaces within $\mathbb{R}^n$. 
A point $\mathcal{Y} \in \mathcal{G}r(k,D)$ is typically represented by an orthonormal basis matrix $Y \in \mathbb R^{n \times k}$ (where $Y^{\top }Y=I_k$). Since any rotation of the basis spans the same subspace, the Grassmannian is intrinsically a quotient space of the Stiefel manifold: $\mathcal{G}r(k, n) = St(k,n)/ \mathcal{O}(k)$. This structure ensures representation invariance to orthogonal transformations (basis rotations). For computation, the manifold is often embedded into the space of projection matrices  $\mathbf{P}=\mathbf{Y}\mathbf{Y}^{\top} $, including projection metric:
\begin{equation}
    d(\mathbf{Y_1}, \mathbf{Y_2})=2^{-1/2}\left \| \mathbf{Y_1}\mathbf{Y_1}^T-\mathbf{Y_2}\mathbf{Y_2}^T \right \|_F, 
\end{equation}
which measures the principal angles between two subspaces.

The properties of the Grassmann manifold offer significant advantages for graph learning. By modeling data as subspaces rather than isolated vectors, it inherently captures high-order structural correlations while filtering out redundant variations. Specifically, its quotient structure ensures that representations remain invariant to basis rotations and node permutations, providing a robust  embedding that is insensitive to noise.


As a novel geometric machine learning method, the Grassmann manifold has been increasingly applied in graph machine learning and related tasks, including graph embedding\cite{yang2021group}, spectral clustering\cite{ma2024cross}, and subspace-based discriminant analysis\cite{wang2024grassmannian}.
Grassmann learning exploits subspace-invariant features and harnesses the structural information, which improves the performance of a model with lower complexity. Grassmann manifolds provide a rigorous space for merging multi-view graph structures. MV-GSL\cite{ghiasi2023multi} and CAGM \cite{ma2024cross} treat different view-specific adjacency matrices as subspaces. By minimizing geodesic distances on the manifold, they fuse multiview information into a unified consensus graph, addressing the inconsistency between graph structures and features without manual hyperparameter tuning. 
Recently,  GyroGr\cite{nguyen2023building} generalized the notions in gyrovector spaces for Grassmann manifolds, and proposed new models and layers for building neural network on Grassmann manifold.


\subsection{Symmetric Positive Definite Manifold}
The Symmetric positive definite (SPD) manifold, $\mathcal{S}^n_{++}$, comprises $n\times n$ symmetric positive definite matrices with strictly positive eigenvalues. It is formally defined as:
\begin{equation}
    \mathcal{S}^n_{++}=\left\{ \mathbf{X} \in \mathbb{R}^{n\times n}| \mathbf{X}=\mathbf{X}^T, \forall \boldsymbol{v} \neq \boldsymbol{0}, \boldsymbol{v}^T \mathbf{X} \boldsymbol{v} > 0 \right \}
\end{equation}
The geometric structure of $\mathcal{S}^n_{++}$ is defined by its Riemannian metric, which dictates its modeling capabilities. Different metrics have been proposed to handle specific data properties, such as Affine-Invariant Metric (AIM) \cite{affinemetric}, Log-Euclidean Metric (LEM) \cite{logmetric}, and Log-Cholesky Metric (LCM) \cite{logcholesky} and and scaling rotation distance\cite{scalingrotaton}. \cite{chen2023adaptive} rethinks LEM and LCM into a general framework, and proposes an adaptive Riemannian metric learned from data. 

Deep learning on SPD manifolds has been widely discussed. Classical methods include the dimensionality reduction algorithm \cite{harandi2016dimensionality}, similarity-based classification \cite{gao2020robusta}, and learning vector quantization \cite{tang2021generalizeda}. 
The field has recently seen significant advances through SPD Neural Networks\cite{huang2016riemannian,dong2017deep,nguyen2019neural,sukthanker2021neural,nguyen2023building,chen2023riemannian,ju2024graph,nguyen2022gyrostructure}, which are deep learning models designed to process SPD metrics directly. To support SPD-based neural network learning, the gradient descent algorithm for the SPD manifold \cite{gao2020learning,han2024riemannian} and normalization techniques \cite{wang2025learning,brooks2019riemanniana} have also been explored. 
SPDNets have proven highly effective in tasks where data is naturally represented by covariance, including face recognition \cite{dong2017deep}, hand gesture recognition \cite{nguyen2019neural}, motor/action imagery classification \cite{ju2024graph,zhang2017deep}, image set classification \cite{wang2022symnet,wang2024spd}, and generative modeling via Denoising Diffusion Probabilistic Models (DDPM) on SPD manifolds for matrix estimation and synthesis\cite{SDPDDPM}.

Although the SPD manifold is rarely applied in graph learning \cite{ju2024graph}, it has powerful application prospects to modeling some key parts in graph learning. For instance, the graph Laplacian is also an element in the SPD manifold, constrained with sparsity. Some graph operations, like graph convolution, can be required to have SPD properties to satisfy somewhat invariance.
In future work, the relationship between the SPD manifold, the graph Laplacian, and some graph operations can be explored more deeply.

\subsection{Generic Manifold}
Unlike previous methods that embed graphs into a predefined space (e.g., hyperbolic), this category treats the graph itself as a discretized manifold. Instead of relying on complex, smooth tensors, it focuses on analyzing the intrinsic geometry of the graph structure. By computing curvature on edges, we can quantify structural properties: positive curvature typically indicates dense connectivity, while negative curvature reveals structural bottlenecks that are responsible for information distortion. 

There are two main discrete versions of Ricci curvature that have been developed. The Ollivier–Ricci curvature is derived from optimal transport considerations. It compares the neighborhood distributions $m_{x}$ and $m_{y}$ around nodes $x$ and $y$ via the Wasserstein distance $W_{1}(m_{x},m_{y})$, and defines curvature as:
\begin{equation}
    \kappa(x,y)=1-\frac{W_{1}(m_{x},m_{y})}{d(x,y)},
\end{equation}
where $d(x,y)$ denotes the graph distance between nodes $x$ and $y$.
In contrast, Forman–Ricci curvature is derived from combinatorial topology. It is computed using simple statistics like node degrees and edge weights . While mathematically simpler and computationally faster than ORC, it effectively captures similar geometric properties.

Since curvature characterizes the contraction and expansion of geometric structures, introducing it into graph representation learning effectively alleviates over-smoothing and over-squashing. For example, SDRF\cite{sdrf} is based on the edge metric Balanced Forman curvature (BFC), finds the edge with the lowest BFC, and proposes a curvature-based rewriting method to alleviate oversquashing.
BORF\cite{nguyen2023revisiting} demonstrated that oversmoothing is linked to positive graph curvature while over-squashing is linked to negative graph curvature, and proposed a graph rewriting algorithm capable of simultaneously mitigating these issues.

\section{Riemannian Neural Architectures} \label{sec: architectures}
\begin{table*}[ht]
    \centering
    \caption{Summary of Riemannian Neural Architectures.}
    \begin{tabular}{ccccccccc}
        \toprule
        Method & Year &Venue & Directed & Dynamic & Heterogeneous& heterophilic & HyperGraph & Architecture  
        \\
        \midrule
        HGCN\cite{hgcn2019nips} & 2019 & NeurIPS & --& --&-- &-- &-- & Convolution 
        \\
        $\kappa$-GCN\cite{Bachmann2020ICML}  & 2020 & ICML &-- &-- & --&-- &-- & Convolution 
        \\
        THGNets\cite{liu2025thgnets}&2025&AAAI&-- &$\checkmark$ &- &-- & $\checkmark$& Convolution
        \\
        LGCN\cite{LGCN21WWW} &2021 & WWW & --& --&-- &-- &-- & Convolution
        \\
        H2H-GCN\cite{h2hgcn2021} &2021 & CVPR & --& --&-- &-- &-- & Convolution
        \\
        TGNN\cite{tgnn2021} &2021 & ACL &-- &$\checkmark$ & --& --& --& VAE \\
         HVGNN\cite{tgnn2021} & 2021& AAAI & --& $\checkmark$&--& --& --& VAE\\
        ACE-HGNN\cite{ace-hgnn2021} & 2021 &ICDM & --& --&-- &-- &-- & Convolution\\
        $\mathcal{Q}$-GCN\cite{Xiong2022NeurIPS} &2022 & NeurIPS &-- & --& --& --& --& Convolution\\
        D-GCN \cite{sun2024aaai} &2024 & AAAI & --&-- & --&--&-- & Convolution\\
        P-VAE\cite{mathieu2019continuous}  &2019 & NeurIPS &-- &-- & --& --& --& VAE\\
        MVAE\cite{skopek2019mixed} & 2020 & ICLR &-- & --&-- &-- &-- &  VAE \\
        HVGAE\cite{liu2025hyperbolic}  &2025 &WWW&$\checkmark$ &-- &-- &-- & --& VAE\\
        H2KGAT\cite{H2KGAT20}& 2020&EMNLP & $\checkmark$ &--& -- & --& --&  Convolution\\
        Hypformer\cite{yang2024hypformer}  & 2024 & KDD & --& --&-- &-- &-- & Transformer\\
        HGCH\cite{HGCH24}&2024&CIKM &-- & --& $\checkmark$&-- & --& Convolution\\
        M2GNN\cite{wang2021mixed}& 2021& WWW & $\checkmark$& --& $\checkmark$ & $\checkmark$& --& Convolution\\
        R-ODE\cite{sun2024r}  &2024 &SIGIR &$\checkmark$ & $\checkmark$&-- & --& --& ODE\\
        Pioneer\cite{Pioneer2025AAAI} &2025 &AAAI &-- & $\checkmark$&-- & --& --& ODE\\
        HGCF\cite{sun2021hgcf}&2021 &WWW & --&-- & $\checkmark$&$\checkmark$ &-- & Convolution\\
        $\mathcal{T C}$-flow &2020 & NeurIPS &-- & --&-- & --& --& flow \\
        GDSS\cite{jo2022score} &2022 &ICML &-- &-- &-- & --&-- & SDE\\
        RicciNet\cite{sunRiccinet}&2024&WWW & --& --&-- &-- &-- & flow model\\
        H2SeqRec\cite{H2SeqRec}& 2021 &CIKM &--&$\checkmark$&$\checkmark$&--&$\checkmark$& Transformer\\
        HGWaveNet\cite{bai2023hgwavenet}&2023 &WWW &$\checkmark$ &-- &-- &-- &--  & Convolution\\
        SHAN\cite{li2023multi} & 2023&CIKM & --& --& $\checkmark$& --& --& Convolution\\
        RotDiff\cite{qiao2023rotdiff}&2023&CIKM&$\checkmark$ &$\checkmark$ & --& --& --& Transformer\\
        HGE\cite{pan2024hge} & 2024 &AAAI &$\checkmark$ &$\checkmark$ & --& --& --& Convolution\\
        Directed-Pseudo\cite{sim2021directed}& 2021&ICML &$\checkmark$ &-- &-- &-- &-- & Convolution\\

        \bottomrule
    \end{tabular}
\end{table*}

\subsection{Riemannian Convolution Networks}

\begin{figure}[htbp]
    \centering
    \includegraphics[width=\linewidth]{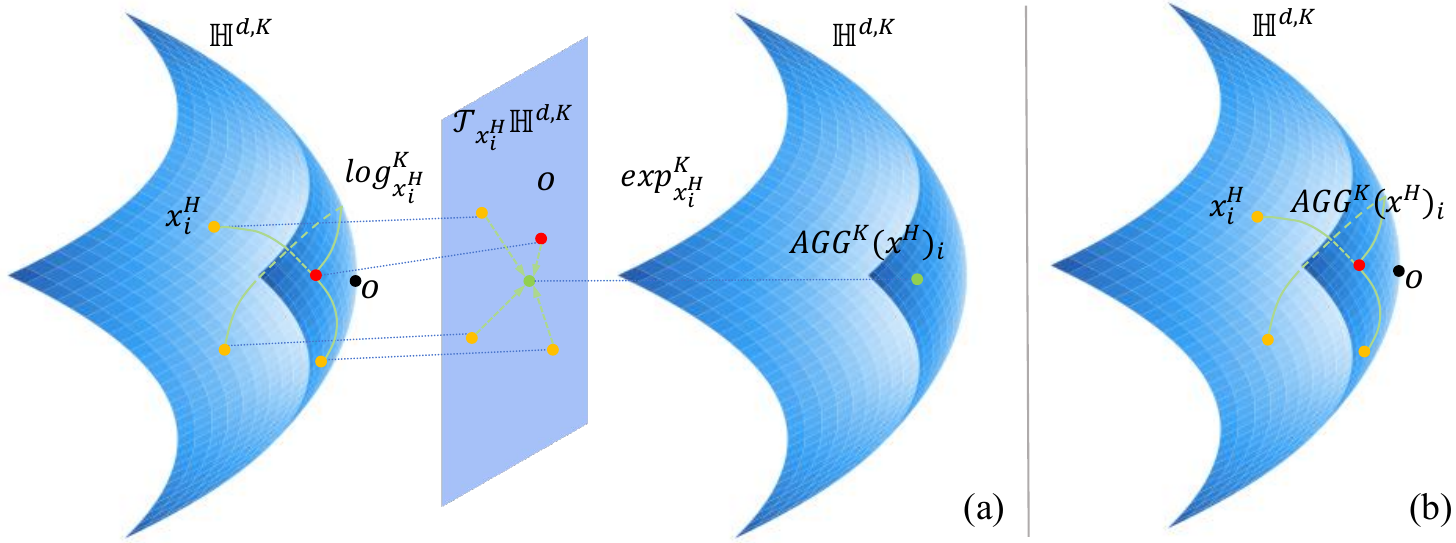}
    \caption{Riemannian Convolution Network Aggregation.}
    \label{fig:gcn}
\end{figure}

\begin{table*}[ht]
    \centering
    \caption{Summary of Riemannian Convolution Networks.}
    \renewcommand{\arraystretch}{2.0} 
    \setlength{\tabcolsep}{6pt}      
    
    \resizebox{\textwidth}{!}{
        \begin{tabular}{l|l|l}
            \toprule
            \textbf{Method} & \textbf{Aggregation} & \textbf{Attentional Weight} \\
            \midrule
            
            \makecell[l]{\textbf{HGCN} \\ \cite{hgcn2019nips}}
            & $\displaystyle \mathbf{x}^H_i= \exp_{\mathbf{x}_i^H}^\kappa \left( \sum_{j \in \mathcal{N}(i)} \alpha_{ij} \log_{\mathbf{x}_i^H}^\kappa (\mathbf{x}_j^H) \right)$ 
            & $\displaystyle \alpha_{ij} = \operatorname{Softmax}_{j} \left( \operatorname{MLP} \left( \log_{\mathbf{o}}^{\kappa} (\mathbf{x}_i^{H}) \,\|\, \log_{\mathbf{o}}^{\kappa} (\mathbf{x}_j^{H}) \right) \right)$ \\
           
            \makecell[l]{\textbf{ACE-HGNN} \\ \cite{ACE-HGNN}}
             & $\displaystyle \mathbf{x}_i = \exp_{\mathbf{h}_{i}^\kappa}^\kappa \left( \sum_{j \in \mathcal{N}(i)} w_{ij} \log_{\mathbf{h}_{i}^\kappa}^\kappa \left( \mathbf{h}_{i\mathbb{H}}^{n,\kappa} \right) \right)$
             & $\displaystyle \alpha_{ij} = \operatorname{softmax}_{j} \left( \operatorname{MLP} \left( \log_{\mathbf{o}}^\kappa ( \mathbf{h}_{i}^\kappa ) \| \log_{\mathbf{o}}^\kappa ( \mathbf{h}_{j}^\kappa ) \right) \right)$ \\
            
            \makecell[l]{\textbf{$\kappa$-GCN} \\ \cite{Bachmann2020ICML}}
            & $\displaystyle \mathbf{x}^\kappa_i = \exp_{\mathbf{x}_i^\kappa}^\kappa \left( \sum_{j \in \mathcal{N}(i)} \alpha_{ij} \log_{\mathbf{x}_i}^\kappa (\mathbf{x}_j) \right)$ 
            & \multicolumn{1}{c}{—} \\
            
            \makecell[l]{\textbf{THGNets} \\ \cite{liu2025thgnets}}
            & $\displaystyle \mathbf{x}_i= \operatorname{proj}_o^\kappa \left( \exp_o^\kappa \left( \sum_{j \in \mathcal{N}(i)} \alpha_{ij} \cdot \log_o^\kappa \mathbf{x}_j \right) \right)$ 
            & $\displaystyle \alpha_{ij} = \frac{\exp(\sigma ( \mathbf{W} \cdot ( \log_o^\kappa \mathbf{x}_i \|\log_o^\kappa \mathbf{x}_j ) ))}{\sum_{k \in \mathcal{N}(i)} \exp(\sigma ( \mathbf{W} \cdot ( \log_o^\kappa \mathbf{x}_i \| \log_o^\kappa \mathbf{x}_k ) ))}$ \\
           
            \makecell[l]{\textbf{LGCN} \\ \cite{LGCN21WWW}}
            & $\displaystyle \mathbf{x}_i= \sqrt{\beta} \frac{\sum_{j \in \mathcal{N}(i) \cup \{i\}} \alpha_{ij} \mathbf{h}_j}{\left\| \sum_{j \in \mathcal{N}(i) \cup \{i\}} \alpha_{ij} \mathbf{h}_j \right\|_L}$ 
            & $\displaystyle \alpha_{ij} = \frac{\exp\left[ -d_L^2 \left( \mathbf{M} \otimes_\beta \mathbf{h}_i, \mathbf{M} \otimes_\beta \mathbf{h}_j \right) \right]}{\sum_{t \in \mathcal{N}(i) \cup \{i\}} \exp\left[ -d_L^2 \left( \mathbf{M} \otimes_\beta \mathbf{h}_i, \mathbf{M} \otimes_\beta \mathbf{h}_t \right) \right]}$ \\
            
            \makecell[l]{\textbf{H2H-GCN} \\ \cite{h2hgcn2021}}
            & $\displaystyle \mathbf{x}_i = p_{\mathcal{B}\to \mathcal{L}}\left(\sigma \left(p_{\mathcal{B}\to \mathcal{L}} \frac{\sum_{j} \alpha_{ij}p_{\mathcal{B}\to \mathcal{L}}(x_j)}{\sum_{j} \alpha_{ij}}\right)\right)$ 
            & $\displaystyle \alpha_{ij}=\frac{1}{\sqrt{1-\left \|h_j^{\kappa}  \right \|^{2} }}$ \\
           
            \makecell[l]{\textbf{$\mathcal{Q}$-GCN} \\ \cite{Xiong2022NeurIPS}}
            & $\displaystyle \mathbf{x}_{i}^{l+1} = \widehat{\exp}^{\kappa}_{\mathbf{o}} \left( \sigma \left( \sum_{j \in \mathcal{N}(i)\cup\{i\}} \widehat{\log}^{\kappa}_{\mathbf{o}} \left( \mathbf{W}^{l} \otimes \mathbf{h}^{l}_{j} \oplus \mathbf{b}^{l} \right) \right) \right)$ 
            & \multicolumn{1}{c}{—} \\
            
            \makecell[l]{\textbf{M2GNN} \\ \cite{wang2021mixed}}
            & $\displaystyle \mathbf{x}_i = \sigma_\kappa \left( \frac{1}{N} \sum_{n, j, k} w^{n}_{ijk} \otimes^\kappa \mathbf{m}^{n}_{ijk} \right)$ 
            & $\displaystyle \alpha_{ij} = \frac{\exp^{\kappa}_{o} ( \text{LReLU} ( \log^{\kappa}_{o} ( \mathbf{W} \otimes^\kappa \mathbf{m}_{ijk} ) ) )}{\sum_{v \in \mathcal{N}(i)} \sum_{q \in \mathcal{R}_{iv}} \exp^{\kappa}_{o} ( \text{LReLU} ( \log^{\kappa}_{o} ( \mathbf{W} \otimes^\kappa \mathbf{m}_{ivq} ) ) )}$ \\
         
            \makecell[l]{\textbf{HGWaveNet} \\ \cite{bai2023hgwavenet}}
            & $\displaystyle \mathbf{x}^{l}_i  = \exp^{\kappa}_{\mathbf{o}} \left( \mathrm{Att} \left( \mathrm{Concat}_{j \in \mathcal{N}(i)} \left( \log^{\kappa}_{\mathbf{o}} ( \mathbf{x}^{l}_{j} ) \right) \right) \right)$ 
            & \multicolumn{1}{c}{—} \\
            \bottomrule
        \end{tabular}
    }
\label{table:neural architectures}
\end{table*}

Graph Convolution Networks (GCNs), a.k.a., Message Passing Neural Networks (MPNNs), follow the message passing paradigm that recursively conducts neighborhood aggregation with graph convolution layer, excelling in modeling the inter-correlation among objects. Without loss of generality, the $k$-th convolution layer takes the form as follows,
\begin{equation}
    \boldsymbol{x}_{u}^{(l)} = \varphi^{(l)} (  \boldsymbol{x}_{u}^{(l-1)}, \operatorname{Agg}( \{  \boldsymbol{x}_{v}^{(l-1)} : (u,v)\in \mathcal E \} ) ),
        \label{eq. mpnn}
\end{equation}
for $k=\{1,\hdots, K\}$, 
where $\operatorname{Agg}$  is a permutation invariant aggregation function, and $\varphi^{(k)}$ denotes the update function of node representation.

Accordingly, aggregation plays a critical role in MPNNs, and we discuss the methods to conduct aggregation on the manifolds. 
\begin{itemize}
    \item \textbf{Tangential Aggregation.} Since it is nontrivial to define a manifold-valued weighted sum, this line of work introduces an additional tangent space to perform feature aggregation. The pair of exponential and logarithmic maps is employed to map between the tangent space and manifold.
    \begin{equation}
    \boldsymbol{x}_{u}^{(l)} =\operatorname{Exp} \left( \varphi^{(l)} \left(  \boldsymbol{x}_{u}^{(l-1)}, \operatorname{Agg} \left( \{ \log (\boldsymbol{x}_{v}^{(l-1)} ) \} \right) \right) \right),
        \label{eq. mpnn-manifold}
\end{equation}
    \item \textbf{Manifold Aggregation.} This lines of work  explicitly defines the manifold-valued weighted sum $\operatorname{Agg}_{\mathcal M}$, and inherits the formulation of traditional MPNN.
\begin{equation}
    \boldsymbol{x}_{u}^{(l)} = \varphi^{(l)} (  \boldsymbol{x}_{u}^{(l-1)}, \operatorname{Agg}_{\mathcal M}( \{  \boldsymbol{x}_{v}^{(l-1)}  \} ) ),
        \label{eq. mpnn-manifold}
\end{equation}
    The manifold-valued weighted sum presents different  formulations according to the model space. Note that these formulations are equivalent in essence, and $\operatorname{Agg}_{\mathcal M}$ typically serves as a key component of fully manifold MPNNs.
\end{itemize}

Graph attentional aggregation is also generalized to the Riemannian counterpart, and key point lies in the definition of attentional weights.
\begin{itemize}
    \item \textbf{Attentional Weight.} It is defined by any scalar function that takes the manifold-value inputs. For instance, the attentional weight can be defined as GAT-type network.
       \begin{equation}
    \alpha(\mathbf{q}_i, \mathbf{k}_j) = \sigma\left( \boldsymbol{\theta}^\top \left[ \mathbf{W} \log^\kappa_{o}(\mathbf{q}_i) \,\|\, \mathbf{W} \log^\kappa_{o}(\mathbf{k}_j) \right] \right),
    \end{equation}
    where $\mathbf{q}_i$ and $\mathbf{k}_i \in \mathcal{M}$ are manifold valued features, $\log^\kappa_{o}(\cdot)$ denotes the logarithmic map at base point $o$ on a constant-curvature manifold with curvature $\kappa$, $\mathbf{W}$  is a learnable weight matrix, and  $\boldsymbol{\theta}$ is a learnable vector for computing attention scores. $\sigma(\cdot) $is a nonlinear activation function.
    
    Also, we can leverage the distance function to derive the attentional weight.
      \begin{equation}
    \alpha(\mathbf{q}_i, \mathbf{k}_j) = \sigma \left( -\beta\, d_{\mathbb{H}}(\mathbf{q}_i, \mathbf{k}_j) - c \right),
    \end{equation}
    where $d_{\mathbb{H}}(\cdot ,\cdot)$ denotes the hyperbolic distance, $\beta$ is a temperature parameter that controls the sharpness of attention, $c$ is a learnable bias.
    
    \end{itemize}

In recent years, a series of Message Passing Neural Networks (MPNNs) have been developed on various non-Euclidean manifolds, including hyperbolic and spherical spaces, among others. 
These models demonstrate promising results in capturing hierarchical, directional, and heterogeneous structures that are difficult to represent in Euclidean space. 
We summarize the formulations of their convolutional layers in Table~\ref{table:neural architectures}.

\textbf{Dynamic Graphs.}  
In addition to embed dynamics data into a manifold space \cite{tgnn2021,LeT24,wang2024ime}, the dynamic perception of structural changes over time can be achieved by modeling the curvature function of time \cite{sun2022self}.
Temporal convolution models curvature as a function of time, enabling smooth and continuous transitions of latent geometry across temporal snapshots. 

\textbf{Heterogeneous Graphs.}
In heterogeneous graphs, recent work ~\cite{msgat2024icdm} proposes assigning independent hyperbolic spaces to different meta paths, where each space is equipped with a learnable negative curvature parameter.

\textbf{Directed Graphs.}
In directed graphs, directional relationships can be effectively captured using hyperbolic distances~\cite{suzuki2019hyperbolic}.

In addition, Riemannian geometry tends to contribute to the mechanistic design of MPNNs.
MPNNs approximate the neural solver of the heat kernel and suffer from a fundamental issue of over-smoothing, which refers to the phenomenon where node representations become indistinguishable when the network goes deep.
Theoretically, incorporating a geometric prior reshapes the functional space with respect to the heat kernel. 
\textbf{GBN} \cite{sundeeper} explores deeper GNNs with Riemannian geometry and addresses the oversmoothing and oversquashing issues in MPNNs, considering the neural scaling law of foundation models.

Existing hyperbolic graph networks often rely on tangent space aggregation, and future work may investigate diverse aggregation schemes that better respect manifold geometry and hierarchical structures.
\subsection{Variational Auto Encoder}

\begin{figure}[htbp]
    \centering
    \includegraphics[width=\linewidth]{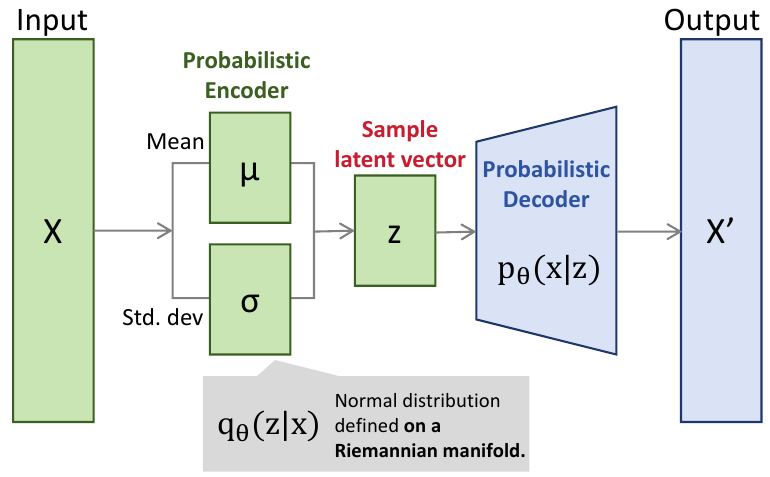}
    \caption{Variational Auto Encoder.}
    \label{fig:gcn}
\end{figure}

Variational autoencoders \cite{kingma2013auto,kingma2019introduction} typically map input data $x$ to a latent Euclidean Gaussian distribution  $  q_\phi(z|x)=\mathcal{N}(\mu,\sigma^2)$. Crucially, they rely on the reparameterization trick for differentiable sampling: 
\begin{equation}
    z = \mu_\phi(x) + \sigma_\phi(x) \odot \epsilon, \quad \epsilon \sim \mathcal{N}(0, I).
\end{equation}

However, this formulation assumes a flat geometry which distorts graph data with intrinsic hierarchies or cycles. Riemannian VAEs replace the Euclidean latent space with a Riemannian manifold. The core challenge lies in defining a valid probabilistic distribution and enabling the reparameterization trick on a manifold, where standard vector addition is undefined .

To model latent variables on manifolds, two primary distribution families are employed: 
\begin{itemize}
    \item \textbf{The Riemannian Normal.} The Riemannian Normal Distribution   \cite{mathieu2019continuous} is defined directly on the manifold using geodesic diatances. 
This distribution is parameterized by the Fréchet mean $\mu$ and a dispersion parameter $\sigma$, and its probability density function is given by:
\begin{equation}
    \mathcal{N}_{\mathcal{M} }( z| \mu, \sigma^2) = \frac{1}{Z(\sigma)} \exp\left(-\frac{d_{\mathcal{M}}(\mu,  z)^2}{2\sigma^2}\right),
\end{equation}
where $d_{\mathcal{M}}$ denotes the induced distance, and $Z(\sigma)$ is the normalization constant ensuring the density integrates to one. 
This formulation allows for a natural generalization of the Euclidean normal distribution to curved spaces, making it suitable for latent-variable modeling in VAEs operating on Riemannian manifolds. 

\item \textbf{The Wrapped Normal.} The Wrapped Normal Distribution   \cite{mathieu2019continuous,nagano1902wrapped} is usually the preferred choice.
It constructs the latent distribution by first sampling a Gaussian vector $v \sim \mathcal{N}(0, \sigma)$ in the tangent space at the origin $\mathcal{T}_o\mathcal{M}$. To center this distribution at a learned mean $u$, the vector is moved to $\mathcal{T}_\mu\mathcal{M}$ via parallel transport $\mathcal{PT}_{0 \rightarrow \mu}(v) $ and then mapped onto the manifold using the exponential map:
\begin{equation}
    z = \mathrm{Exp}_\mu\left( \mathcal{PT}_{0 \rightarrow \mu}(v) \right), \quad v \sim \mathcal{N}(0, \sigma).
\end{equation}
This mechanism effectively wraps the flat noise onto the curved surface, enabling a mathematically rigorous yet differentiable extension of the reparameterization trick.

\end{itemize}

Several works extend Variational Autoencoders (VAEs) to hyperbolic spaces to capture hierarchical structure~\cite{mathieu2019continuous,park2021unsupervised,cho2023hyperbolic}.
Mixed-curvature latent spaces combining Euclidean, hyperbolic, and spherical components have been proposed to improve flexibility in modeling diverse data geometries~\cite{skopek2019mixed}.
More general Riemannian formulations leverage intrinsic manifold priors such as Brownian motion or SPD submanifolds to enable geometry-aware inference with uncertainty modeling~\cite{miolane2020learning,kalatzis2020variational}.

The Riemannian Normal is theoretically the most principled, using geodesic distance and satisfying the maximum entropy property. 
However, it is computationally expensive due to intractable normalization and rejection-based sampling.
The Wrapped Normal offers a good trade-off, combining efficient sampling, reparameterizability, and support for full covariance. Though it lacks a maximum entropy interpretation, it is well-suited for deep learning applications such as VAEs on manifolds.

\subsection{Riemannian Transformer}

Unlike Graph Attention Networks (GATs), which restrict the self-attention mechanism to a node's immediate neighbors as defined by the graph structure, Graph Transformers capture global dependencies via the self-attention mechanism. 
This relies on two core operations. The first is attention score, calculated via the scaled dot product of query and key vectors:
\begin{equation}
    \alpha_{i,j} \propto (W_Q h_i)^{\top} (W_K h_j)
\end{equation}
The second is feature aggregation, performed as a weighted sum of value vectors: \begin{equation}
    h'_i = \sum_{j} \mathrm{softmax}(\alpha_{i,j})(W_V h_j)
\end{equation}
However, these operations, such as inner products similarity and vector addition for aggregation, are geometrically ill-defined on curved manifolds.

Riemannian Transformers \cite{ermolov2022hyperbolic,huang2024combinatorial,spinner2024lorentz} enable the modeling of complex geometric structures in data. 
Recent works have focused on adapting the self-attention mechanism to Riemannian spaces \cite{gulcehre2018hyperbolic,HGAT2021,yang2024hypformer}, particularly hyperbolic geometry, to capture hierarchical relationships in data.
A key insight behind these advancements lies in extending classical linear operators to Riemannian manifolds.
Sun et al.~\cite{sun2023self} proposed a linear operation in adaptive Riemannian spaces.
A Riemannian feature transformation is defined through the \(\kappa\)-left-multiplication operator, denoted by \(\otimes_{\kappa}\). 
This operator leverages the manifold's exponential and logarithmic maps to enable geometry-preserving transformations.
Formally, given an input point \(\mathbf{h} \in \mathcal{M}^{d,\kappa}\) and a weight matrix \(\mathbf{W} \in \mathbb{R}^{d' \times d}\), the Riemannian linear transformation is defined as:
\begin{equation}
\mathbf{W} \otimes_{\kappa} \mathbf{h} := \exp^{\kappa}_{\mathcal{O}} \left( \left[ 0 \middle\| \mathbf{W} \log^{\kappa}_{\mathcal{O}}(\mathbf{h})[1{:}d] \right] \right),
\label{eq:r-linear}
\end{equation}
where \(\exp^{\kappa}_{\mathcal{O}}(\cdot)\) and \(\log^{\kappa}_{\mathcal{O}}(\cdot)\) denote the exponential and logarithmic maps at the origin point \(\mathcal{O}\) on the manifold \(\mathcal{M}^{d,\kappa}\), and \([\cdot \| \cdot]\) indicates vector concatenation. The notation \(\log^{\kappa}_{\mathcal{O}}(\mathbf{h})[1{:}d]\) refers to the first \(d\) coordinates of the tangent vector, while the zero-padding ensures that \([\log^{\kappa}_{\mathcal{O}}(\mathbf{h})]_0 = 0\) holds for all \(\mathbf{h} \in \mathcal{M}^{d,\kappa}\).

Wang et al. \cite{LGCN21WWW}  designed canonical transformations in the hyperbolic space to avoid violating the principles of hyperbolic geometry, the formulation is:
\begin{equation}
    f(\mathbf{x}) := \exp_0^\kappa \bigl( f\bigl( \log_0^\kappa(\mathbf{x}) \bigr) \bigr), \hat{f}(\mathbf{v}) := \bigl( 0, f(v_1, \cdots, v_n) \bigr).
\end{equation}
This method only leverages the Euclidean transformations on the last $n$ coordinates $(v_1, \cdots, v_n)$ in the tangent space, and the first coordinate $v_0$ is set to 0 to satisfy the constraint.
Furthermore, Yang et al.~\cite{yang2024hypformer} introduced a graph transformer named HypFormer, which operates on the hyperbolic manifold and can be viewed as a special case of the above Riemannian transformation.
Riemann Transformers show clear advantages in tasks with hierarchical, symmetric, or geometrically structured data. 
These models effectively integrate the core strengths of traditional Transformer architectures, which excel at capturing long-range global dependencies through self-attention mechanisms, with geometric inductive biases that respect the underlying data manifold structure.
Future work should aim to optimize the trade-off between computational efficiency and representational capacity to further enhance the scalability and applicability of Riemann Transformers in large-scale, complex geometric learning tasks.

\subsection{Riemannian Graph Ordinary Differential Equations}

\begin{figure}[htbp]
    \centering
    \includegraphics[width=\linewidth]{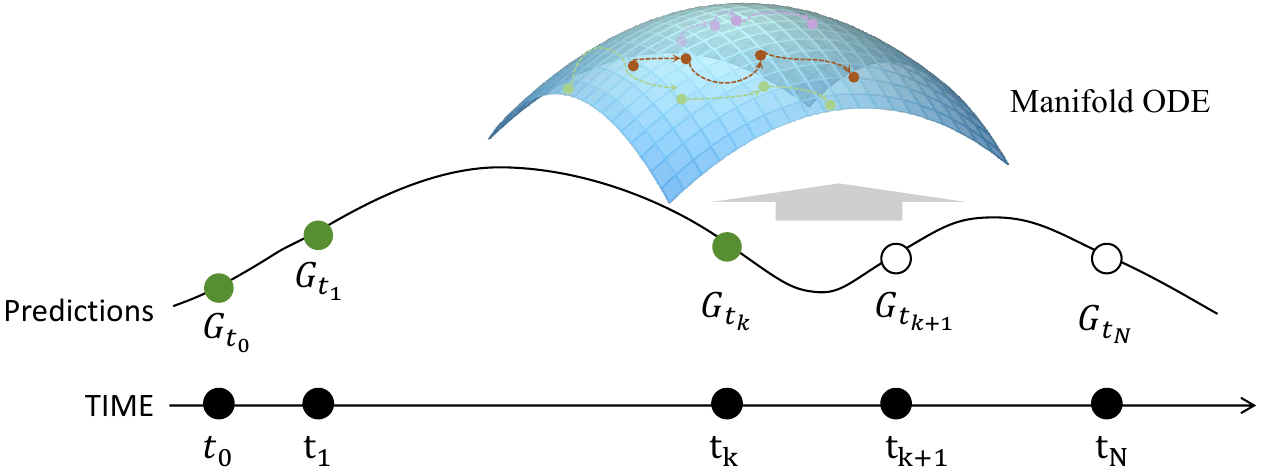}
    \caption{Riemannian Graph Ordinary Differential Equation.}
    \label{fig:ode}
\end{figure}

Graph Neural Ordinary Differential Equations (ODEs) have been proven effective for modeling dynamics interacting system.
The trajectory of objects are described by ODEs, whose velocity (or  vector field) is parameterized by graph neural networks.
The formulation of graph neural ODEs is typically written as follows: 
\begin{equation}
    \frac{d\mathbf{H}(t)}{dt}=\text{GNN}(\mathbf{H}(t),\mathcal{G};\Theta).
\end{equation}

Recent years have witnessed a surge of extending to Riemannian manifolds.
Specifically, the objects move on the manifold $h_t \in \mathcal{M}$, and the vector field resides in the tangent bundle $ \frac{d \mathbf{H}(t)}{dt} \in \mathcal{TM}$ accordingly. 
That is, it requires a tangent space-valued vector field that takes inputs of manifold points.

Next, researchers consider the physic law underlying the dynamics: energy and entropy.

Hamiltonian mechanics provides an elegant framework to study the dynamics of both conservative and dissipative (energy) systems.
Itcharacterizese the dynamic systems with an ODE defined on a Riemannian manifold, e.g., smooth symplectic manifolds are able to encode the system's intrinsic geometric properties. 
In Hamiltonian mechanics, the Hamiltonian $\mathcal{H}(q,p)$ is a function of the generalized coordinate $q$ and canonical momentum $p$. The dynamics of the system follow Hamilton's equations, which are a pair of first-order ODEs taking the form as follows,
\begin{align}
    \dot{q} \equiv  \frac{dq}{dt} = \frac{\partial \mathcal{H}}{\partial p}, \quad \dot{p} \equiv  \frac{dp}{dt} = -\frac{\partial \mathcal{H}}{\partial q},
\end{align}
where the Hamiltonian vector field $\mathcal{X}_{\mathcal{H}}=(\frac{\partial \mathcal{H}}{\partial p}, -\frac{\partial \mathcal{H}}{\partial q} )$.

There are some researchers have explored the Hamiltonian physics laws using neural networks, such as HNN\cite{hamiltonianneural}, HOGN\cite{HOGN}, CDGH\cite{CDGH}.

Recent advances in geometric analyses have shown the connection between system entropy and Ricci curvature. 
In the literature, the Ricci flow like constraint is introduce to regulate the entropy pattern in the system evolvement, e.g., aligning with entropy increasing principle thermodynamics. 

 In more general dynamical systems, Euclidean-based approaches often neglect the intrinsic geometry of the system and physics laws, e.g., the principle of entropy increasing. $\textit{Pioneer}$\cite{Pioneer2025AAAI} proposed a novel physics-informed Riemannian graph ODE for a wider range of entropy-increasing dynamic system, which ensures geometry-aware and physics trajectories law. The entropy-increasing is achieved by a $\textit{Constrained Ricci Flow}$, as shown:
 \begin{equation}
     \frac{d w_{ij}^{t}}{dt} = (Ric_{ij}^{F} - e^{f(z_{i}^{t},z_{j}^{t};\Theta)})w_{ij}^{t},
 \end{equation}
 where $f_{\Theta}$ is a parameterized model to constrain the canonical Ricci flow, correcting the curvature evolution to ensure consistency with the entropy increase principle.

The aforementioned efforts primarily describe the physics-informed dynamics with first-order ODE systems.
Integrating Riemannian manifolds with second-order ODEs has potential to model more sophisticated dynamics while preserving the intrinsic geometry.  
We also note that the efficiency of simulating ODEs limits model performance.

\subsection{Riemannian Denoising Diffusion and SDE}

\begin{figure}[htbp]
    \centering
    \includegraphics[width=\linewidth]{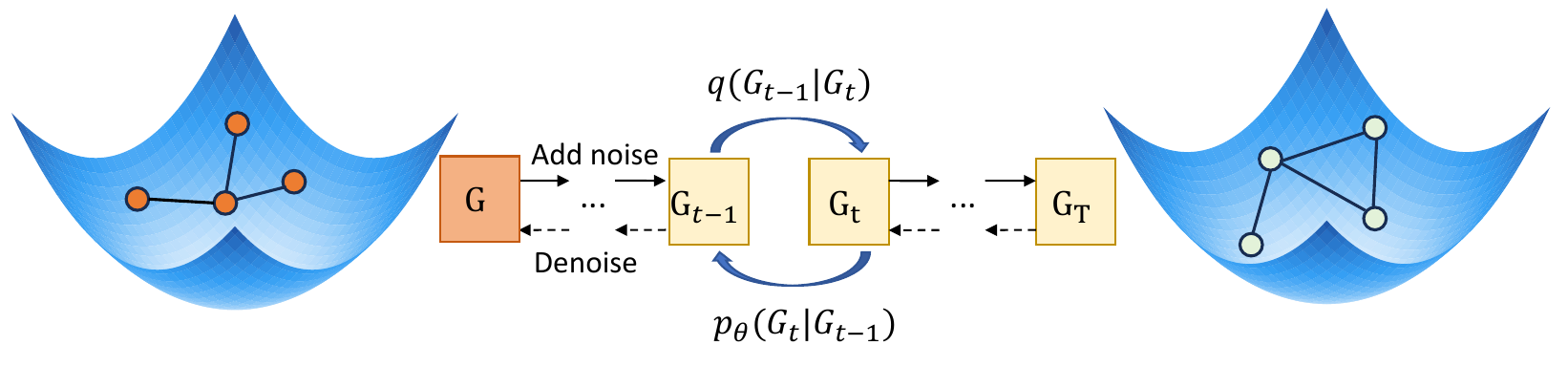}
    \caption{Riemannian Denoising Diffusion and SDE.}
    \label{fig:sde}
\end{figure}

Denoising diffusion models generate  data by learning to reverse the gradual process that corrupts data with noise.
In Euclidean space, this framework is naturally modeled by Stochastic Differential Equations (SDEs). The forward SDE slowly injects noise into data $x_0$ o transform it into a prior distribution (e.g., Gaussian) over time $t$:
\begin{equation}
    \mathrm{d}\mathbf{x} = \mathbf{f}(\mathbf{x}, t) \mathrm{d}t + g(t) \mathrm{d}\bar{\mathbf{w}},
\end{equation}
where  $\mathbf{f}(\cdot, t)$ is the drift coefficient of $\mathbf{x}(t)$, $g(\cdot)$ is the diffusion coefficient, and $\mathbf{w} $ denotes a standard Brownian motion.
The generation process is achieved by solving the reverse time SDE:
\begin{equation}
    \mathrm{d}\mathbf{x} = \left[ \mathbf{f}(\mathbf{x}, t) - g(t)^2 \nabla_{\mathbf{x}} \log p_t(\mathbf{x}) \right] \mathrm{d}t + g(t) \mathrm{d}\bar{\mathbf{w}},
\end{equation}
where $\nabla_{\mathbf{x}} \log p(\mathbf{x})$ is known as the score function.

Extending this framework to a Riemannian manifold $\mathcal{M}$ requires replacing Euclidean Brownian motion with Riemannian Brownian motion $B_{t}^{\mathcal{M}}$ and defining the drift using intrinsic geometry. The geometric SDE is formulated as: 
\begin{equation}
    dX_{t}=- \frac{1}{2}\nabla _{X_{t}}U(X_{t})dt+dB_{t}^{\mathcal{M}},
\end{equation}
where the terminal distribution satisfies $\mathrm{d}p(x)/\mathrm{dvol}_x \propto e^{-U(x)} $ for a potential function $U$.
However, implementing the reverse process on manifolds faces a significant bottleneck: the calculation of the heat kernel $p_{t|0}(x_t|x_0)$. Unlike in Euclidean space where the heat kernel is a simple Gaussian, on manifolds it typically lacks a closed-form expression, making the exact computation of the score function $\nabla_{\mathbf{x}} \log p(\mathbf{x})$ intractable. 

To overcome these geometric obstacles, researchers have developed solutions evolving from approximation to variational inference: 

(1) Approximate Sampling: RSGM \cite{de2022riemannian} employed Geodesic Random Walks (GRWs) to simulate the SDE, where noise is sampled in the tangent space and mapped back to the manifold via the exponential map,
 \begin{equation}
    X_{n+1}^{\gamma} = \exp_{X_{n}^{\gamma}}[\gamma f(n \gamma, X_{n}^{\gamma})+\sqrt{\gamma}\sigma(n\gamma,X_{n}^{\gamma})Z_{n+1}],
\end{equation}
where $\gamma$ is the step-size, $Z_{n+1}$ is the Gaussian in the tangent space of $X_{n+1}^{\gamma}$. Moreover, to approximate the score function,  RSGM leverages Varadhan’s asymptotic formula, approximating the intractable heat kernel gradient with the gradient of the geodesic distance for small time steps:
\begin{equation}
    \lim _{t \rightarrow 0} t \nabla_{x_{t}} \log p_{t \mid 0}\left(x_{t} \mid x_{0}\right)=\exp _{x_{t}}^{-1}\left(x_{0}\right).
\end{equation}

(2) Variational Inference: RDM\cite{HuangABPC22} introduced a more general variational learning framework that bypasses explicit heat kernel computation. It implicitly performs score matching by maximizing a Riemannian continuous-time ELBO, using the Projected Hutchinson Trace Estimator to efficiently compute the divergence of vector fields.

These principled frameworks for modeling data on manifolds have found significant applications, such as molecular generation, and earth and climate science.  For molecular generation, HypDiff\cite{HypDiff24} proposed a hyperbolic geometric latent diffusion process that is constrained by both radial and angular geometric properties, ensuring the preservation of the original topological properties in the generative graphs. HGDM\cite{HGDM24AAAI} proposed a two-stage hyperbolic graph diffusion model to better learn the distribution of graph data. For the earth and climate science, RSGM\cite{de2022riemannian}, RDM\cite{HuangABPC22} and LogBM\cite{icmlJoH24} constructed various frameworks to simulate these non-Euclidean data on manifolds.

\subsection{Riemannian Flow Model and Flow Matching}

\begin{figure}[htbp]
    \centering
    \includegraphics[width=\linewidth]{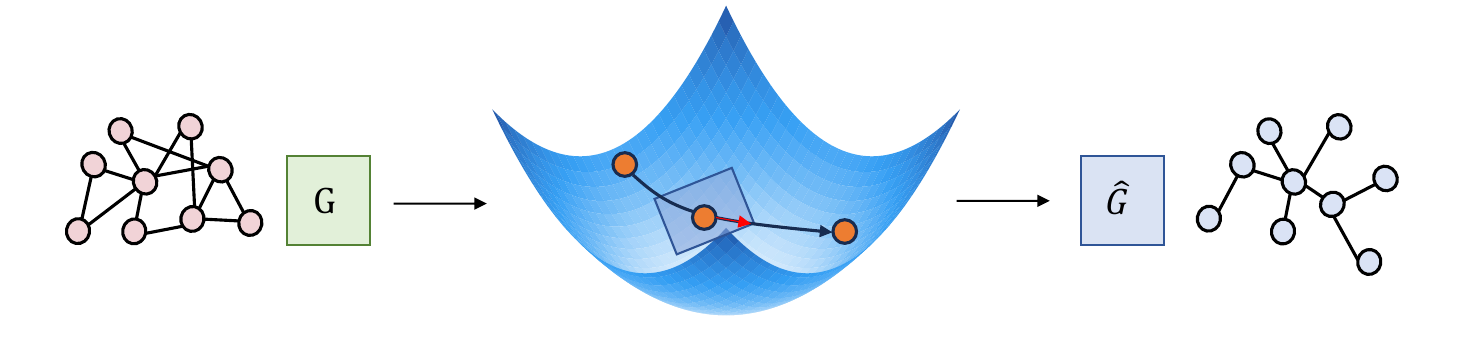}
    \caption{Riemannian Flow Model and Flow Matching.}
    \label{fig:flow}
\end{figure}

Riemannian flow models originally consider the Continuous Normalizing Flows (CNFs), which describe generative processes using Ordinary Differential Equations (ODEs). On a manifold $\mathcal{M}$, the trajectory of a point evolves under a time-dependent vector field $v_t$ on the tangent space. Early works extended this framework to specific geometries, such as Tori and Spherical CNFs \cite{rezende2020normalizing} and Hyperbolic CNFs \cite{BoseSLPH20}. However, the training process requires intensive ODE integration and divergence evaluation, which prevents scalability to high-dimensional or complex geometries. 

To overcome these limitations and provide a common foundation for both ODE flows and SDE diffusion models, the perspective of Riemannian measure space is adopted. Here, generative modeling is reframed as learning probability measures on the measure space $\mathcal P(C[0,1], \mathcal M^{\kappa, d})$. This high-level perspective shifts the objective from ``solving an equation" to ``matching a trajectory distribution". 

\subsubsection{Flow Matching}
Flow matching considers a probabilistic flow that maps a simple distribution (e.g., Gaussian distribution) to a complex data distribution of interest. It enables simulation-free training by directly regressing the vector field $v_t$ to match a set of constructed conditional paths. 

On manifolds, Riemannian flow matching \cite{DavisKPCBB24} leverages geodesics as natural target paths between a pair of samples $(x_0, x_1)$ from the distribution $p(x_0)$ and the target distribution $q(x_1)$,
\begin{equation}
    x_t  = \operatorname{exp}_{x_1}^{\kappa}(t \operatorname{log}_{x_1}^{\kappa}(x_0)), \quad t \in [0,1].
\end{equation}
The loss function aims to minimize the discrepancy between the model's predicted vector field $v_t$ and the target vector filed,
\begin{equation}
    \mathcal{L}(\theta)=\mathbb{E}_{t, q\left(x_{1}\right), p\left(x_{0}\right)}\left\|v_{t}\left(x_{t}\right)-\dot{x}_{t}\right\|_{g}^{2}
\end{equation}
where $\dot{x}_{t} = \frac{d}{dt} x_t$ and $\left \| \cdot \right \| _g$ is the norm induced by the Riemannian metric $g$. This approach completely avoids ODE integration and divergence computation, enabling scalable training on complex manifolds.

\subsubsection{Bridge Matching}
As a more generic case, Schr\"odinger bridge models \cite{shi2023diffusion,liu2022deep,gushchin2024light} the mapping between two distributions, i.e., source distribution and target distribution. 
It aims to find the process that minimizes the KL divergence to a reference Wiener process 
  \begin{equation}
    T^{\star} = \arg \min \left \{ KL(T||W):T_0 = p_0, T_1=p_1 \right \}, 
\end{equation}
where $W$ is a reference process, typically set as the Wiener process of Brownian motion, and the optimal process $T^{\star}$ is referred to as the Schr\"odinger bridge. 
The classic methods to solve the Schr\"odinger bridge problem, such as Iterative Proportional Fitting (IPF)\cite{IPF} and Iterative Markovian Fitting (IMF)\cite{IMF23nips}, which involve iterative optimizations that are computationally expensive and can result in error accumulation.
Trace\cite{sun2025trace} introduced a new algorithm built upon Riemannian optimal projection. The Riemannian Schr\"odinger family $\mathcal{S}$ collects the stochastic process of Schr\"odinger bridge $S$, pinned at $p_0$ and $p_1$ in the Riemannian measure space. The Riemannian reciprocal process is written as $T_{\pi}=\operatorname{proj}_{R^{\mathcal{M}}}(T)=\int _{\mathcal{M} \times \mathcal{M}}W_{|x_0,x_1}^{\mathcal{M}}d\pi(x_0,x_1)$ with Riemannian Wiener process.  Riemannian optimal projection of $T_{\pi}$ leads to Riemannian Schr\"odinger Bridge $T^{\star}$ between the distribution $p_0$ and $p_1$,
\begin{equation}
    T^{\star}= \operatorname{proj}_{s^{\mathcal{M}}}(T_{\pi})=\operatorname{arg}_{S\in \mathcal{S}}\operatorname{min}(T_{\pi}||S).
\end{equation}
This converts the SB problem into a single-step regression problem, significantly improving scalability and stability.

\section{Learning Paradigms}\label{sec: paradigms}
\begin{figure*}[htbp]
    \centering
    \includegraphics[width=1\linewidth]{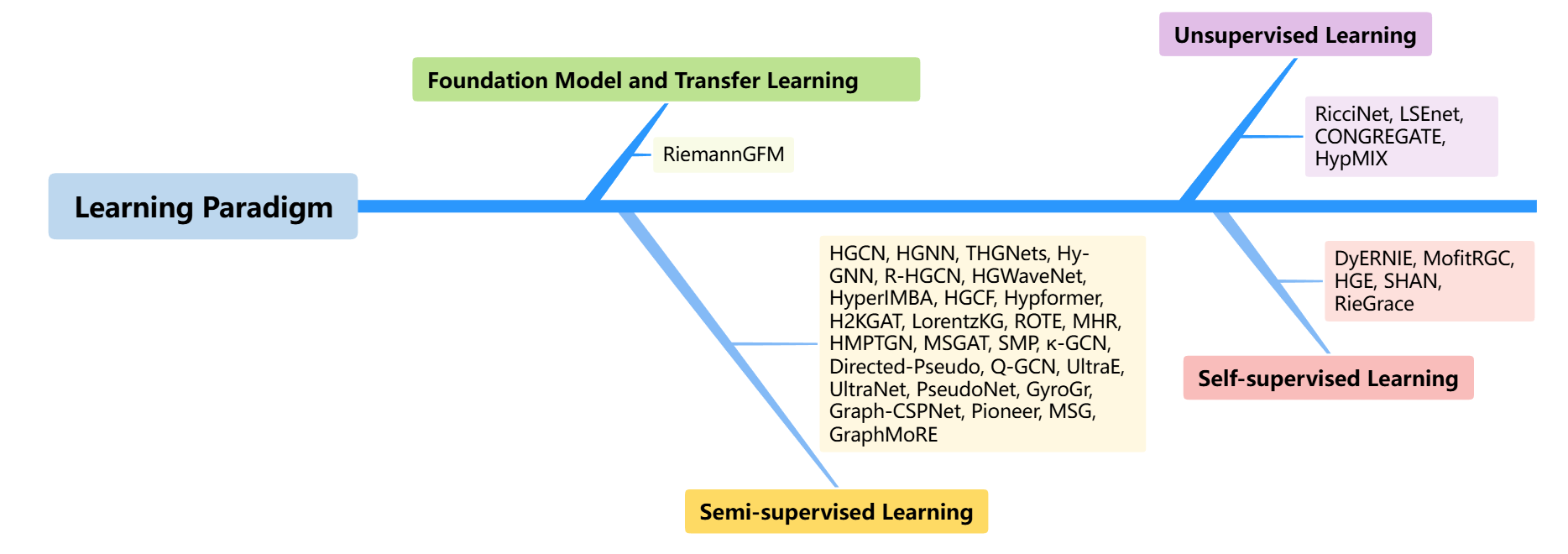}
    \caption{Summary of Learning Paradigms.}
    \label{fig:placeholder}
\end{figure*}
\subsection{Semi-supervised Learning}

Semi-supervised learning on graphs exploits structural information to enhance learning performance when labeled data is limited. Many traditional graph tasks (e.g., node classification, link prediction, and graph classification) naturally fall under the semi-supervised setting. In Euclidean models, structural mismatches can lead to distorted propagation. Riemannian spaces enhance this paradigm by treating the manifold's geometry as a strong inductive bias. 
For instance, embedding a social network into hyperbolic space naturally separates high-degree root nodes from leaf nodes. This geometric alignment acts as a regularizer, ensuring that labels propagate correctly along the intrinsic hierarchy, leading to superior classification performance even with limited supervision.

A key challenge is mapping node features from different tangent spaces onto a common manifold coordinate system to enable meaningful aggregation.
In Riemannian geometry, the exponential map and logarithmic map serve as fundamental tools for this purpose, enabling transformations between the manifold and its associated tangent space to facilitate effective feature alignment.

For a given point $x \in \mathcal{M}$ and a tangent vector $v \in T_x\mathcal{M}$, there exists a unique geodesic $\gamma: [0,1] \rightarrow \mathcal{M}$ such that $\gamma(0) = x$ and $\gamma'(0) = v$. 
The \textbf{exponential map} at point $x$, denoted $\exp_x: T_x\mathcal{M} \rightarrow \mathcal{M}$, is defined as
\begin{equation}
\exp_x(v) = \gamma(1),
\label{eq:exp_map}
\end{equation}
which projects vectors from the tangent space to the manifold along geodesics. 
The exponential map allows features to be placed onto the curved manifold while preserving the geodesic structure, ensuring that subsequent computations respect the manifold’s geometry.

Its inverse, the \textbf{logarithmic map} $\log_x: \mathcal{M} \rightarrow T_x\mathcal{M}$, maps manifold points back to the tangent space, enabling linear operations in a locally Euclidean neighborhood:
\begin{equation}
\log_x(y) = \gamma'(0), \quad \text{where } \gamma(1) = y.
\label{eq:log_map}
\end{equation}

These maps enable Riemannian GNNs to perform feature transformations in tangent space while respecting the global manifold structure during aggregation and propagation.

During model training, the input features are first mapped onto the manifolds via the exponential map. For instance, these features could be produced by pre-trained Euclidean neural networks.
To perform learnable transformations, the manifold features are then always projected back to the tangent space at a reference point through the logarithmic map, where standard linear operations can be safely applied. 
Non-linear transformations in the tangent space, such as Riemannian versions of activation functions (e.g., ReLU, LeakyReLU, or hyperbolic-specific activations), are then employed to introduce model expressiveness while respecting the underlying geometry. 
Finally, the updated tangent vectors are mapped back to the manifold via the exponential map before being propagated through the graph.
\subsection{Unsupervised Learning}
Unsupervised learning on graphs, clustering, aims to understand the structures in the graph without explicit labels. In Euclidean space, deep graph clustering methods first learn informative node representations and then perform traditional clustering algorithms, such as K-Means and DBSCAN.

However, traditional Euclidean clustering often fails to capture complex intrinsic topologies. Riemannian methods address this by aligning the embedding space with the graph's structure. For example, \textsc{Congregate}\cite{sunCONGREGATE} considered graph clustering from a geometric respective, and constructed a heterogeneous curvature space where representations are generated via the product of Riemannian graph convolutional nets. 
Additionally, the inherent hierarchy of hyperbolic spaces naturally expresses the clustering organization of graphs. A recent advance aligns the structural entropy to the hyperbolic expression of graph clustering\cite{sunLSEnet}. Structural entropy is a principled establishment for encoding graph organization from the information theory. However, it has not been introduced to cluster and gaps roots in the discrete formulation before.
Optimal encoding tree can be regarded as a discrete analogy to the hyperbolic space. Accordingly, LSEnet\cite{sunLSEnet}, models the encoding tree in the hyperbolic space and interplays the hyperbolic representations and tree construction. It receives significant performance gain against the original structural entropy. A promising direction is to explore the geometric information on the manifold, e.g., hyperbolic structural entropy for clustering.

Beyond static embeddings, Riemannian geometry introduces a dynamic clustering paradigm: Ricci Flow. This process treats the graph as a manifold that evolves over time to reveal its community structure. 

The evolution is governed by the partial differential equation:
\begin{equation}
    \frac{\partial }{\partial t}d_t(x,y) = -d_{t}(x,y)\operatorname{Ric}(x,y), \label{eq:ricciflow}
\end{equation}
Ricci flow is a dynamic geometric process that divides a smooth manifold into different regions according to the graph structure. The dynamic process is governed by Ricci curvature. The regions of large positive curvature shrink in whereas regions of very negative curvature spread out.

Traditional Ricci curvature is discrete and non-differentiable. RicciNet \cite{sunRiccinet} formulated a differentiable Ricci curvature parameterized by a Riemannian graph convolution. It feeds data points collected to Riemannian graph convolution on the Laplacian $L^{\alpha}$, 
\begin{equation}
    Y^{\alpha} = \delta(L^{\alpha}(X \otimes_{\kappa}W)),
\end{equation}
where $\otimes_{\kappa}$ denotes $\kappa$-right multiplication, $W$ is the weight matrix. Then, Ricci curvature is derived as follows,
\begin{equation}
    \operatorname{Ric}^{\alpha}(i,j)=1-(\left \|y_i  \right \|_0 -\left \| y_j \right \|_0  )/d(x_i,x_j).
\end{equation}
Crucially, it enforces a geometric regularity. The distance between node pairs during the decoding process is trained to match the theoretical solution of the Eq. \ref{eq:ricciflow} . This constraint forces the neural network to simulate the geometric flow, effectively separate the graph into distinct clusters along geodesic paths.

\subsection{Self-supervised Learning}
Self-supervised learning (SSL) aims to learn informative representations by exploring the similarities within the data themselves.
A representative method is known as contrastive learning, which learns from an augmented view where the mutual information is maximized.
Unlike the vision domain, where augmentation is easily obtained by rotation and flipping, graph augmentation is challenging since the methods of node dropout and edge perturbation disrupt structural regularity.

Riemannian manifolds provide the potential of augmentation-free contrastive learning. 
In the literature, \cite{sun2024aaai} introduces geometric graph contrastive learning.
Different geometric spaces (e.g., Euclidean, hyperbolic, and spherical) capture complementary aspects of graph structure, offering intrinsic diversity for contrastive objectives. 
The core challenge of geometric contrastive learning lies in achieving agreement between positive and negative samples - the similarity measure between the Riemannian manifold of different curvatures. 

There exist two main strategies for aligning the two geometric views.

One is to use a common tangent space, in which contrastive learning is performed at the north pole $o$ of the manifold \cite{RGFM25sun}. 
Let \( z_i^H \in \mathcal{T}_{p_i^H} \mathcal{H} \) and \( z_i^S \in \mathcal{T}_{p_i^S} \mathcal{S} \) denote the tangent-space encodings of node \( i \) in the hyperbolic and spherical manifolds, respectively. To compare these embeddings directly, we parallel transport both to a shared tangent space at \( o \), yielding \( PT_{p_i^H \rightarrow o}(z_i^H) \) and \( PT_{p_i^S \rightarrow o}(z_i^S) \). The geometric contrastive loss is then defined as:

\begin{equation}
    \begin{split}
        \mathcal{J}(H, S)  = & \\
    - \sum_{i=1}^N \log \frac{\exp\left( \left\langle PT_{p_i^H \rightarrow o}(z_i^H),; PT_{p_i^S \rightarrow o}(z_i^S) \right\rangle \right)}{\sum_{j=1}^N \exp\left( \left\langle PT_{p_i^H \rightarrow o}(z_i^H),; PT_{p_j^S \rightarrow o}(z_j^S) \right\rangle \right)},
    \end{split}
\end{equation}

where \( \langle \cdot, \cdot \rangle \) denotes the standard Euclidean inner product in the shared tangent space \( T_o \mathcal{M} \). This formulation allows embeddings from different manifolds to be directly compared in a geometry-neutral space, effectively removing curvature-induced incompatibilities.

This construction, known as the \textbf{product of tangent bundles} over composite manifolds, enables full expressiveness in modeling both intrinsic node attributes and extrinsic relational geometry. 
Importantly, it facilitates \textbf{feature transfer} across the graph by transporting tangent features between nodes via \textbf{parallel transport}, 
\begin{equation}
PT_{p_i^\mathcal{M} \rightarrow \mathbf{o}}(z_i^\mathcal{M}),
\end{equation}
where $\mathbf{o} \in \mathcal{M}$ is a predefined reference point (e.g., the origin). 
The operator denotes the parallel transport along the geodesic from $\mathbf{P}_i^\mathcal{M}$ to $\mathbf{o}$, enabling the alignment of tangent vectors in a common coordinate system $\mathcal{T}_{\mathbf{o}}\mathcal{M}$ for comparison and aggregation.

Another strategy is to map one view onto the other, thereby enabling a direct comparison in the same manifold space.
Formally, given a node $i$, let $\mathbf{z}_i^m$ be the geometric view and $\mathbf{z}_i^o$ be the Euclidean view. 
The loss is defined as: 
\begin{equation}
\mathcal{L} = - \sum_{i=1}^{N} \log \frac{\exp\left(s\left(\mathbf{z}_i^m, \mathbf{z}_i^0\right)\right)}{\sum_{j=1}^{N} \exp\left(h\left(\mathbf{z}_i^m, \mathbf{z}_j^0 \mid \mathcal{M}_3\right) \cdot s\left(\mathbf{z}_i^m, \mathbf{z}_j^0\right)\right)},
\label{eq:rc-motif}
\end{equation}
where $s$ is a hyperparameter, and $h(\cdot,\cdot|\mathcal{M}_3)$ is a motif-aware hardness weighting function, modulating the influence of each negative sample $j$.
It reduces to standard InfoNCE when the hardness weighting function is uniform, i.e., when $h(\cdot) = 1 $. 
For example, RieGrace\cite{sun2023self} proposed Generalized Lorentz Projection(GLP) to map the manifolds with different dimensions and curvatures. Formally, it designed a transformation matrix $\mathbf{W} \in \mathbb {R}^{d_2 \times d_1}$, and the $( GLP_{\mathbf{x}}^{d_1, \kappa_1 \rightarrow d_2, \kappa_2}(\cdot) )$ is defined as:
\begin{equation}
    GLP_{\mathbf{x}}^{d_1, \kappa_1 \rightarrow d_2, \kappa_2} \left( \begin{bmatrix} w & \mathbf{0}^\top \\ \mathbf{0} & \mathbf{W} \end{bmatrix} \right) \begin{bmatrix} x_0 \\ \mathbf{x}_s \end{bmatrix} = \begin{bmatrix} w_0 x_0 \\ \mathbf{W} \mathbf{x}_s \end{bmatrix},
\end{equation}
where $w \in \mathbb{R}$, $w_0 = \sqrt{\frac{|\kappa_1|}{|\kappa_2|} \cdot \frac{1 - \kappa_2 \ell(\mathbf{W}, \mathbf{x}_s)}{1 - \kappa_1 \langle \mathbf{x}_s, \mathbf{x}_s \rangle}}$.

In summary, Riemannian contrastive learning leverages the inherent geometric diversity of multiple manifolds to provide augmentation-free, complementary views.
It enables more expressive and structure-aware graph representations.

\subsection{Foundation Model and Transfer Learning}

Foundation models, addressing numerous learning tasks with a universal pretrained model, have marked a revolutionary advancement such as Large Language Models (LLMs)\cite{achiam2023gpt}.
Unlike the success in the language and vision domains, foundation models are still in their infancy for learning on graphs.
As traditional GNNs are typically tuned by specific tasks, Graph Foundation Models (GFMs) have gathered increasing research attention recently.
GFMs refer to a novel family of graph neural networks that are pre-trained on broad graph data at scale, empowering transferability among the graphs or tasks. 
Recent advances show the potential of Riemannian geometry to address the graph transferability.
 Some recent advances leverage the LLMs to build GFMs, while the sequential graph description tailored for the language model tends to fail in capturing the structural complexity.
 Another line of work introduces the graph prompting techniques for GNNs to improve the transferability.

Analogous to foundation models in NLP and CV, GFT\cite{GFT24} proposed that GFMs may rely on a universal vocabulary that encodes transferable patterns shared across different tasks and domains.
GFT considered tree structures derived from the message-passing process as the fundamental unit for transferable vocabulary, which preserves the essential graph structural information.
The choice of computation trees over graph motifs is motivated by the inherent limitations of basic GNNs in identifying certain motifs, such as cycles, which restrict their capacity to capture richer structural patterns. The core of GFT's pre-training phase lies in learning a universal graph vocabulary through a computation tree reconstruction task. Specifically, they adopted Vector Quantization (VQ) to discretize the continuous embedding space of computation trees into a set of
tokens. The pre-training objective function consists of three main components:
\begin{equation}
\mathcal{L}_{tree}=\beta_1 \cdot \mathcal{L}_{feat} + \beta_2 \cdot \mathcal{L}_{sem} + \beta_3 \cdot \mathcal{L}_{topo},
\end{equation}
where $\mathcal{L}_{feat}$ reconstructs the features of the root node, $\mathcal{L}_{sem}$ reconstructs the overall semantics of the computation trees, and $\mathcal{L}_{topo}$ reconstructs the connectivity among nodes in the computation trees.

Recently, Riemannian geometry has provided a new paradigm to fundamentally address the challenge of graph foundation models.
Specifically, Riemannian geometry offers a construction to model the common substructures in the graph domain for universal model pre-training-- the product bundle.
The rationale lies in that the geometry and attributes of certain substructures can be expressed in a factor bundle, and the shared substructures among different graphs are utilized for knowledge transfer.
Sun et al. \cite{RGFM25sun} chose a simple yet effective set of common substructures: trees and cycles.
Given the alignment between hyperbolic space and trees (spherical space and cycles), the product bundle is constructed as follows, 
\begin{equation}
    \mathcal{P}^{dp} = \mathcal{H}_{\kappa_H}^{d_H} \otimes T\mathcal{H}_{\kappa_H}^{d_H} \otimes \mathcal{S}_{\kappa_S}^{d_S} \otimes T\mathcal{S}_{\kappa_S}^{d_S},
\end{equation}
where \( \otimes \) denotes the Cartesian product and \( d \) and \( \kappa \) are the dimensions and curvatures of the respective manifolds. 
Accordingly, it adopts the disentangled representation among structure and attribute information, and the node \( \mathbf{x}_i \) encoding is written as 
\begin{equation}
    \mathbf{x}_i = [\mathbf{p}_i^H \,|\, \mathbf{z}_i^H \,|\, \mathbf{p}_i^S \,|\, \mathbf{z}_i^S],
\end{equation}
where \( \mathbf{p}_i^H \in \mathcal{H}_{\kappa_H}^{d_H} \) and \( \mathbf{z}_i^H \in T_{\mathbf{p}_i^H} \mathcal{H}_{\kappa_H}^{d_H} \) are the structure and attribute encoding in the hyperbolic factor, 
respectively, while  \( \mathbf{p}_i^S \in \mathcal{S}_{\kappa_S}^{d_S} \) and \( \mathbf{z}_i^S \in T_{\mathbf{p}_i^S} \mathcal{S}_{\kappa_S}^{d_S} \) are those quantities in  the hyperspherical counterpart. 
With the manifolds instantiated by $\kappa$-stereographical models, 
Riemannian metric of the product bundle is yielded  as
\begin{equation}
     {g}^{\mathcal{P}}_x =  {g}^{\kappa^H}_x \oplus I_{d_H+1} \oplus  {g}^{\kappa^S}_x \oplus I_{d_H+1}
\end{equation}
where $I_{d_H+1}$ is the $(d_H + 1)$ dimensional identity matrix, and $\oplus$ denotes the direct sum among  matrices.

A promising future direction lies in unifying diverse graph domains into a coherent Riemannian manifold through topological-manifold fusion, which could provide a systematic differential geometry framework for understanding knowledge transfer and establishing geometric scaling laws in graph foundation models.

\section{Datasets and Open Sources} \label{sec: datasets}

We summarize commonly used datasets or benchmarks for Graph in Table \ref{table:benchmark_datasets}. 
We observe that different research applications use different datasets and benchmarks.
In the context of benchmarking data sets for graph-based machine learning tasks, the selected datasets span multiple domains, including citation networks, biochemical graphs, social networks, and other diverse graph structures. These datasets vary significantly in scale and complexity, ranging from small-scale networks such as Cora with 2708 nodes to massive graphs like Reddit with over 232,000 nodes and more than 116 million edges. They differ in structural characteristics—such as average number of nodes, edges, features, and classes—and serve a wide range of research purposes, including node classification, link prediction, and graph clustering. 
\begin{table*}[htbp]
\centering
\caption{Summary of selected benchmark data sets.}
\label{table:benchmark_datasets}
\begin{tabularx}{\textwidth}{p{1.5cm}|llllllll}
\toprule
\multicolumn{1}{l}{Category} & \multicolumn{1}{l}{Data set} & \multicolumn{1}{l}{Source} & \multicolumn{1}{l}{\# Graphs} & \multicolumn{1}{l}{\# Nodes(Avg.)} & \multicolumn{1}{l}{\# Edges (Avg.)} & \multicolumn{1}{l}{\#Features} & \multicolumn{1}{l}{\# Classes} & \multicolumn{1}{l}{Citation} \\
\midrule
\multirow{4}{*}{\rotatebox[origin=c]{0}{\parbox{2cm}{Citation\\Networks}}} & Cora &{\cite{Cora-Citeseer}} & 1 & 2708 & 5429 & 1433 & 7 & {\cite{sunCONGREGATE}}, {\cite{sun2024aaai}}, {\cite{GraphMoRE}},{\cite{Bachmann2020ICML}},\\
& & & & & & & & 
{\cite{law2021ultrahyperbolic}}, {\cite{Xiong2022NeurIPS}}, {\cite{sunLSEnet}}, {\cite{yang2024hypformer}}, \\
& & & & & & & & 
{\cite{RHGCNXue2024AAAI}}, {\cite{hgcn2019nips}}, {\cite{HyperIMBA23}}, {\cite{MalikGK25}} \\
\cmidrule(r{5pt}){2-9}
 & Citeseer & {\cite{Cora-Citeseer}} & 1 & 3327 & 4732 & 3703 & 6 & {\cite{sunCONGREGATE}}, {\cite{sun2024aaai}}, {\cite{GraphMoRE}}, {\cite{RGFM25sun}}, \\
& & & & & & & & 
{\cite{Bachmann2020ICML}},{\cite{law2021ultrahyperbolic}}, {\cite{Xiong2022NeurIPS}}, {\cite{sunLSEnet}},\\
& & & & & & & & 
{\cite{yang2024hypformer}}, {\cite{RHGCNXue2024AAAI}}, {\cite{HyperIMBA23}} \\
\cmidrule(r{5pt}){2-9}
 & Pubmed & {\cite{Pubmed}} & 1 & 19717 & 44338 & 500 & 3 & {\cite{sun2024aaai}}, {\cite{GraphMoRE}}, {\cite{RGFM25sun}}, {\cite{Bachmann2020ICML}},  \\
& & & & & & & & 
{\cite{law2021ultrahyperbolic}}, {\cite{Xiong2022NeurIPS}}, {\cite{yang2024hypformer}}, {\cite{RHGCNXue2024AAAI}},  \\
& & & & & & & & 
{\cite{hgcn2019nips}} \\
\cmidrule(r{5pt}){2-9}
 & DBLP & {\cite{DBLP}} & 1 & 17716 & 105734 & 1639 & 4 & {\cite{tgnn2021}}, {\cite{msgat2024icdm}}, {\cite{HypMix24}} \\
\midrule
\multirow{3}{*}{\rotatebox[origin=c]{0}{\parbox{2cm}{Bio-chemical\\Graphs}}} & PPI & {\cite{PPI}} & 24 & 56944 & 818716 & 50 & 121 & {\cite{hgcn2019nips}}\\
\cmidrule(r{5pt}){2-9}
 & Enzymes & {\cite{Enzymes}} & 600 & ~32.6 & ~124.3 & 7 & 2 & {\cite{law2021ultrahyperbolic}} \\
\cmidrule(r{5pt}){2-9}
 & QM9 & {\cite{QM9}} & 133885 & ~18.0 & ~37.3 & 11 & 19 & {\cite{yataka2023grassmann}}, {\cite{smp22iclr}}, {\cite{hnnnips19}} \\
\midrule
\multirow{2}{*}{\rotatebox[origin=c]{0}{\parbox{2cm}{Social\\Networks}}} & Reddit & {\cite{Reddit}} & 1 & 232965 & 11606919 & 602 & 41 & {\cite{tgnn2021}}\\
\cmidrule(r{6pt}){2-9}
 & Actor & {\cite{Actor}} & 1 & 7600 & 33544 & 931 & 5 & {\cite{HyperIMBA23}} \\
\midrule
\multirow{11}{*}{\rotatebox[origin=c]{0}{\parbox{2cm}{Others}}} & MNIST & {\cite{MNIST}} & 70000 & ~70.6 & ~564.5 & 3 & 10 & {\cite{mohammadi2024generalized}}, {\cite{BoseSLPH20}} \\
\cmidrule(r{5pt}){2-9}
 & Airport & {\cite{hgcn2019nips}} & 1 & 3188 & 18631 & 4 & 4 & {\cite{sun2024aaai}}, {\cite{RGFM25sun}}, {\cite{Bachmann2020ICML}}, {\cite{Xiong2022NeurIPS}},  \\
& & & & & & & & 
{\cite{yang2024hypformer}}, {\cite{RHGCNXue2024AAAI}}, {\cite{hgcn2019nips}}, {\cite{MalikGK25}} \\
 \cmidrule(r{5pt}){2-9}
 & AFEW & {\cite{AFEW}} & - & - & - & - & - & {\cite{sukthanker2021neural}}, {\cite{huang2016riemannian}} \\
 \cmidrule(r{5pt}){2-9}
 & Chameleon & {\cite{Chameleon}} & - & 2277 & 31421 & - & 5 & {\cite{HyperIMBA23}} \\
 \cmidrule(r{5pt}){2-9}
 & Computer & {\cite{Pitfall18}} & 1 & 13752 & 491722 & 767 & 10 & {\cite{MSG2024nips}}, {\cite{sunLSEnet}} \\
 \cmidrule(r{5pt}){2-9}
 & FB15k-237  & {\cite{WN18RR}} & - & 14951 & 1345 & - & - & {\cite{xiong2022ultrahyperbolic}}, {\cite{LorentzKG}}, {\cite{ATTH20}},  \\
& & & & & & & & 
{\cite{FFHR}} \\
 \cmidrule(r{5pt}){2-9}
 & HDM05 & {\cite{HDM05}} & - & - & - & - & - & {\cite{sukthanker2021neural}}, {\cite{zhang2017deep}}, {\cite{huang2016riemannian}}, {\cite{GyroGr}} \\
 \cmidrule(r{5pt}){2-9}
 & Photo & {\cite{Pitfall18}} & 1 & 7650 & 238162 & 745 & 8 & {\cite{MSG2024nips}}, {\cite{sunCONGREGATE}}, {\cite{GraphMoRE}}, {\cite{sunLSEnet}},  \\
& & & & & & & & 
{\cite{HyperIMBA23}} \\
 \cmidrule(r{5pt}){2-9}
 & SKIG & {\cite{SKIG}} & - & - & - & - & - & {\cite{piao2019double}}, {\cite{wang2018cascaded}} \\
 \cmidrule(r{5pt}){2-9}
 & WN18RR  & {\cite{WN18RR}} & - & 40943 & 18 & - & - & {\cite{xiong2022ultrahyperbolic}}, {\cite{LorentzKG}}, {\cite{ATTH20}}, \\
& & & & & & & & 
{\cite{FFHR}} \\
 \cmidrule(r{5pt}){2-9}
 & YAGO3-10 & {\cite{YAGO3-10}} & - & - & - & - & - & {\cite{xiong2022ultrahyperbolic}}, {\cite{ATTH20}} \\
\bottomrule
\end{tabularx}
\end{table*}

\section{Practical Applications}  \label{sec: applications}

\begin{figure*}[htbp]
    \centering
    \includegraphics[width=1\linewidth]{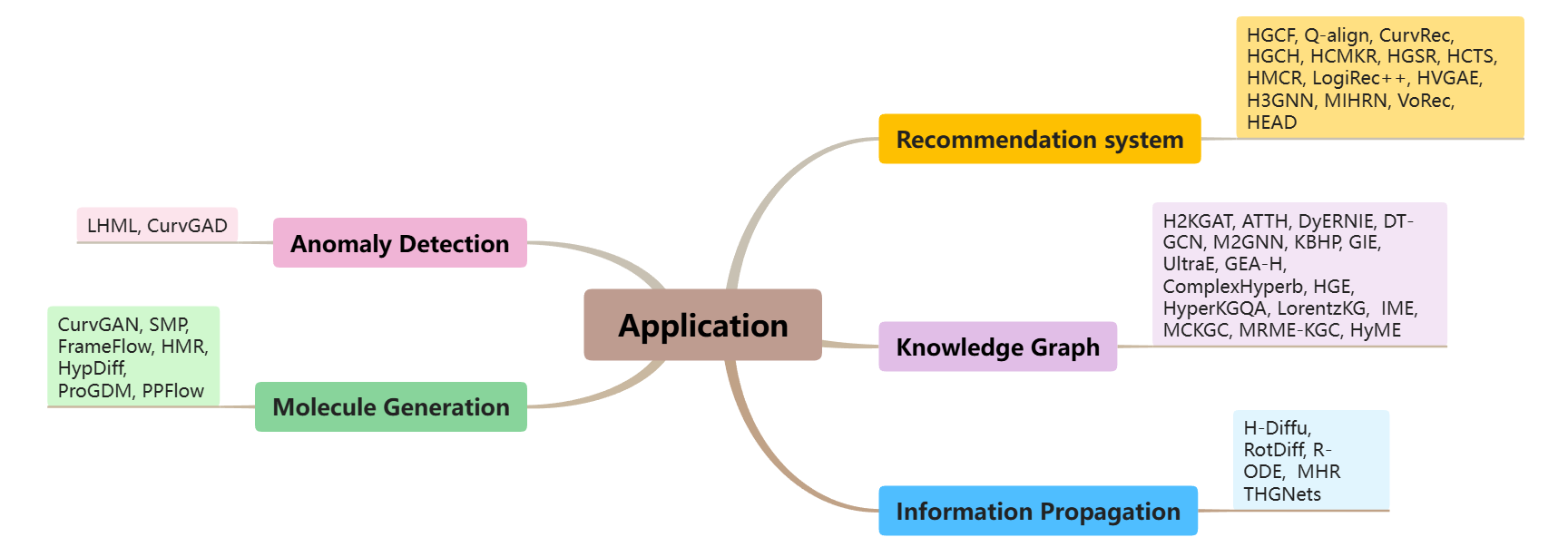}
    \caption{Summary of Applications.}
    \label{fig:placeholder}
\end{figure*}

Riemannian Graph Learning has been applied to various disciplines.
In this section, we summarize existing applications into five key fields.
\subsection{Recommender Systems}

Graph-based recommender systems have gained increasing attention for their ability to model user–item interactions as complex networks, enabling high-quality personalized recommendations. Its core challenge is that interaction graphs exhibit a power-law distribution, strongly implying a natural hierarchical structure. Traditional Euclidean GCNs squash this tree-like structure onto a flat space, causing severe representation distortion. 
Hyperbolic space, with its exponential volume growth, is an ideal geometry for embedding such data with low distortion. 
Several representative studies have incorporated Riemannian principles into recommendation models. HGCF \cite{sun2021hgcf} introduces hyperbolic geometry to collaborative filtering, preserving hierarchical structures through tangential convolution and capturing higher-order user–item associations via multi-level aggregation. H2SeqRec \cite{H2SeqRec} extends this idea to sequential recommendations by embedding temporal behavior hierarchies in hyperbolic space. To address label sparsity, HyperCL \cite{HyperCL} performs contrastive learning between the hyperbolic manifold and its tangent space, while LKGR \cite{LKGR} employs the Lorentz model to encode hierarchical associations and mitigate sparse interactions. 
There are also some works focus on enhancing the quality representations through improved manifold operations. FHGCN \cite{FHGCN} mitigates the projection distortion with tangent spaces through the fully hyperbolic convolution, and HGCH \cite{HGCH24} considers the power law prior in hyperbolic gated fusion, while adopting negative sampling to speed up convergence. To handle heterogeneity and dynamic interactions, HGSR \cite{HGSR} tackles the network heterogeneity with hyperbolic social embedding pre-training and alleviates the noise in user preference learning, and SINCERE \cite{SINCERE}, considers the sequential interaction network and introduces the co-evolving Riemannian manifolds to address the complex dynamics between users and items. 
While most current research focuses on hyperbolic spaces, extending Riemannian graph learning to other manifold geometries could further enhance generalization and adaptability.

\subsection{Knowledge Graph}

Knowledge Graphs (KGs) in Euclidean space pose fundamental challenges.  Many KGs contain inherent non-Euclidean structures, such as hierarchies. Euclidean geometry suffers from high representation distortion when embedding these structures in low dimensions. This has motivated the use of Riemannian manifolds as a more appropriate geometric inductive bias. 
Early efforts focused on leveraging the tree-like properties of pure hyperbolic spaces to model hierarchies. 
For instance, 
ATTH \cite{ATTH20} introduced hyperbolic isometries (e.g., rotations, reflections) with attention in the Poincaré ball model to capture complex logical patterns in low dimensions.  
LorentzKG \cite{LorentzKG} represents entities as points in the Lorentz model and represents relations as intrinsic transformations to overcome distortions and inaccuracies. 
However, real-world KGs are heterogeneous, containing hierarchical, cyclic, and flat structures concurrently. A single, fixed curvature is insufficient. This limitation led to mixed-curvature spaces. 
M2GNN\cite{wang2021mixed} pioneered embedding KGs into a product space of multiple single-curvature manifolds  
Along a similar trajectory, T-GCN \cite{DT-GCN21} defines a continuous curvature space and fully embeds attribute values of data types, reducing the omission of structural information.
The next frontier for static KG modeling is Temporal Knowledge Graphs,  which introduce the dimension of time to capture the evolution of facts. 
IME \cite{wang2024ime} modeled the temporal knowledge graph in multi-curvature spaces to capture complex geometric structures. 
Similarly, HGE\cite{pan2024hge} extended the temporal knowledge graph to the product space, embedding temporal facts in different subspaces via two temporal attention mechanisms. Beyond modeling specific structures, ensuring the quality of the embeddings is important. 
The application of Riemannian geometry to KGs shows a clear trajectory. Despite significant progress, computational complexity, the interpretability of mixed spaces, and the automatic selection of optimal geometric combinations for specific tasks remain critical open questions for future research.

\subsection{Molecule Generation}

In the domain of molecular generation, researchers usually model the molecules as graphs where atoms and chemical bonds are denoted as nodes and edges respectively. 
A fundamental challenge is that any generated molecule must satisfy intrinsic physical principles, such as translation and rotation equivariance (SE(3)-equivariance). 
While molecules are realized in 3D Euclidean space, their latent space of graphs is itself highly non-Euclidean and anisotropic. 
Applying generative models directly in a Euclidean latent space can lead to the loss of topological information. 
Applying diffusion models directly in a Euclidean latent space can lead to the loss of topological information. 
For instance, HypDiff \cite{HypDiff24} performed discrete graph diffusion under the guidance of topological properties based on hyperbolic space to improve model quality. 
HGDM \cite{HGDM24AAAI}  captured the crucial graph structure distributions by constructing a hyperbolic potential node space that incorporates edge information, thereby reducing distortions and interference in modeling the distribution. 
ProGDM \cite{ProGDM24} proposed a graph diffusion model approach based on mixed-curvature product spaces. 
These methods primarily leverage Riemannian geometry to optimize the latent topological distribution of graphs. 
The more formidable challenge is generating a molecule's 3D conformation, which demands strict adherence to SE(3)-equivariance. Riemannian geometry offers two distinct solutions for this.
One approach bypasses the SE(3) problem by modeling on Torus Manifolds. Since torsion angles are inherently rotation-invariant, PPFlow\cite{pptflow} proposed a conditional flow-matching model on torus manifolds to learn the distribution of peptide backbone torsion angles directly. 
A more prominent approach models directly on the SE(3) manifold itself. 
FrameDiff\cite{FrameDiff} developed an SE(3)-invariant diffusion model for protein modeling, equipped with an SE(3)-equivariant score network that does not need to be pretrained. 
FrameFlow\cite{frameflow} adapted FrameDiff to the flow-matching generative modeling paradigm and reduced the cost compared to previous methods. 
FoldFlow\cite{FOLDFLOW} utilized SE(3) flow matching and further introduced variants using Riemannian Optimal Transport (OT) to learn more stable flows. 
While generative models have significantly improved sampling efficiency, future challenges remain. These include the efficient generation of larger, more complex biomolecules, the tighter integration of geometric frameworks with accurate side-chain packing, and the exploration of novel manifolds (e.g., molecular surfaces ) to capture specific physicochemical properties.

\subsection{Anomaly Detection}
Graph anomaly detection aims to identify nodes exhibiting behaviors or characteristics deviating from normal. 
Traditional Graph Anomaly Detection (GAD) focuses on structural or attribute deviations, but Riemannian geometry provides a novel perspective by identifying geometric outliers. 
DNAD-RCA \cite{DNAD-RCA} proposed a  Riemannian-based optimization, which addresses the requirement for centralized observations in the low-rank matrix completion problem.
On the representation front, early explorations leveraged the power of specific manifolds. 
HNN-GAD\cite{hnngad} combined the hyperbolic space for better graph representations and better separation of normal nodes and outliers. The hyperbolic space could split normal nodes and outliers with large margins.
More recent works explore deeper into local geometric information, particularly curvature, to enhance the interpretability of anomaly detection. 
\textbf{CurvGAD} \cite{CurvGAD} introduced an innovative dual-pipeline framework. The first pipeline is curvature-equivariant geometry reconstruction, which learns representations in mixed-curvature spaces and reconstructs the graph's curvature matrix. The second pipeline is curvature-invariant structure and attribute reconstruction. It reconstructs the adjacency and feature matrix while ensure that the process remains invariant to the curvature. The dual-pipeline architecture captures diverse anomalies across homophilic and heterophilic networks.
Future directions may include exploring more efficient curvature estimation algorithms and extending these geometric insights to dynamic graphs and complex graph-level anomaly detection tasks.

\subsection{Information Propagation}
Information propagation is fundamental to understanding the structure and underlying dynamics of social networks. 
Given the complexity of underlying diffusion rules, researchers have turned to Riemannian geometry, which offers distinct tools to model the dynamics of the process and the representation of complex structure.
Riemannian geometry was first applied to more accurately represent the underlying structure of propagation networks. Some works, like H-Diffu \cite{FengZFFWLS23}, use hyperbolic space to capture the hierarchy of user influence. This concept was extended to multi-modal contexts, such as fake news detection, where MHR\cite{feng2025mhr} employs a Lorentzian fusion module and learns the graph representation in hyperbolic space. Info-HGCN \cite{Info-HGCN} employed the hyperbolic graph convolution network to capture the social hierarchy of users, addressing the issue of ineffectively modeling the social hierarchy of users. 
However, information propagation involves more than just static hierarchies. 
THGNets \cite{liu2025thgnets} introduced a hyperbolic temporal hypergraphs neural network to alleviate the distortion of user features by hyperbolic hierarchical learning in information cascades. 
RotDiff \cite{QiaoFLLH0Y23} utilized rotation transformations in hyperbolic space to model complex diffusion patterns, addressing the issues of asymmetric features of the diffusion process. Rather than merely representing the network's static hierarchy, a separate branch focuses on modeling the dynamic process of propagation itself.  R-ODE \cite{sun2024r} leveraged this insight by incorporating Ricci curvature to guide a continuous-time dynamic model controlled by a Neural Ordinary Differential Equation (Neural ODE). This enables precise, time-aware predictions of personalized information diffusion pathways.

\section{Future Direction}  \label{sec: future}

In this section, we analyze existing challenges in Riemannian graph learning, and highlight $8$ promising directions for future research.

\subsection{Benchmarking}
Riemannian graph learning has garnered significant interest in recent years, with numerous methods developed for diverse domains including chemistry, physics, social sciences, knowledge graphs and recommendation systems. Developing powerful and expressive models has been a key concern for advancing the practical applications so far. However, tracking progress in this field remains challenging due to the lack of benchmarks. Models are often evaluated on traditionally used datasets with inconsistent experimental comparisons, making it difficult to effectively distinguish the performance differences between architectures.
Existing GNN benchmarks are insufficient for a comprehensive study of Riemannian graph learning.
Future research should focus on building large-scale, diverse benchmark datasets that cover a wide range of real-world applications. These datasets need to include graphs with different topologies, curvatures, and sizes. Additionally, developing unified evaluation metrics and protocols is crucial.

\subsection{Graph Diversity}
Existing studies on Riemannian graph learning primarily work with undirected, homogeneous graphs. However, real-world graphs can be directed, heterophilic, or heterogeneous. For example, social networks have asymmetric relationships (e.g., follower and followee). Recommender systems consist of a heterogeneous collection of users, items and comments.
Biological proteins show heterophilic connections between functionally distinct nodes. Thus, novel methods are needed to model the edge directions, capture heterophilic patterns, and encode heterogeneous node/edge features. 
Correspondingly, the features of Riemannian manifolds are still underexplored, e.g., asymmetric operations in gyrovetor space, and manifold functional analysis for long-range dependence.

\subsection{Model Expressiveness}
So far, the Riemannian manifolds other than hyperbolic spaces are less visited, and off-the-shelf operations are still limited. The repeated usage of the exponential and logarithmic maps tends to introduce numerical errors, while the parameterization of the important notions such as Riemannian connections and cohomology largely remains open. We also note that recent efforts primarily focus on formulation generalization, rather than on model or mechanism adjustment with geometric priors. That is, the superior expressiveness of Riemannian geometry has not been fully unleashed. Taking MPNNs as an example, a series of Riemannian variants are defined on different manifolds, and have achieved encouraging results due to better structural alignment.
Most existing studies generalize the Euclidean formulation to their manifold counterparts, and thereby the fundamental issues such as over-smoothing and over-squashing, are inherited by these Riemannian models. 
Thus, there is still a need for new message passing paradigms that are empowered to address these systematic shortcomings with advanced geometric tools.

\subsection{Model Architecture}
As mentioned above, Riemannian models have shown initial success in graph convolution networks, variational autoencoders, neural ODEs, SDEs, and flow matching. However, the exploration of Riemannian models remains in its infancy given the wide family of deep neural architectures. For instance, Riemannian geometry has rarely been applied in reinforcement learning (RL).
RL thrives in robotic control (manipulation, navigation), autonomous driving (lane changes, collision avoidance), game AI (Go), and resource allocation. Riemannian metrics enable distance measurements for nonlinear state/action spaces like robotic $SO(3)$ joints; geodesics stabilize high-dimensional optimization (e.g., autonomous driving trajectories).
These strengths make Riemannian geometry a promising tool with untapped potential for diverse deep neural architectures.

\subsection{Model Interpretation}
Interpretability has become a critical focus in deep learning research, as the opacity of complex models impedes trust, validation processes, and real-world deployment. Interpretation in graph learning presents an equally pressing challenge.
Take autonomous driving systems as an illustration: while graph learning techniques can process sensory inputs to make real-time decisions, the specific reasoning processes through which these decisions are made and the underlying rationales remain crucial; engineers must know why the model chooses to accelerate, brake, or change lanes to ensure safety.

However, a deeper exploration of Riemannian geometry in graph learning may provide novel insights into this issue.
For instance, geometric perspectives offer unique insights into how data points behave on manifolds: by examining how data moves along geodesics, we can better understand models capture relationships between different data instances. Additionally, the geometric properties like curvature and local tangent spaces directly influence how graph learning models propagate information and update node representations.

\subsection{Model Scalability}
Scalability is critical for graph learning models to adapt as real-world graph data evolves.
As graph structured data expands across domains, from massive social networks to intricate molecular graphs, both the scale (node/edge volume) and complexity (heterogeneous relationships, rich attributes) of graphs surge. 
For graph models, performance depends not only on capturing intrinsic graph patterns but also on maintaining or enhancing such performance as data volume and model size grow. 
Traditional models often falter here: constrained by fixed architectures or inefficient computation, they may stagnate or decline when faced with larger, more complex graphs.
This drives an urgent demand for scalable Riemannian graph learning solutions. 
However, existing Riemannian learning methods that rely on extensive exponential/logarithmic mappings are computationally expensive and hinder the scalability for large graphs. 
Future research should focus on efficient architectures that capture Riemannian manifold intrinsic properties (conformality, equivariance, symmetry). 

\subsection{Riemannian Graph Foundation Model (R-GFM)}
Foundation models have achieved remarkable success in various fields (e.g., large language models). 
It is interesting to build graph foundation models (GFMs) with both expressiveness and universality. 
One line of GFMs considers text-attributed graphs; their transferability depends on textual commonality, tending to generate suboptimal results on graphs without textual attributes. 
GFMs built with Large Language Models (LLMs) use sequential graph descriptions and fail to capture the core characteristic of graphs -- structural complexity. Additionally, existing efforts are mainly confined to Euclidean spaces. GFMs are still far from practical reality, especially for structural transferability.

Riemannian geometry empowers graph foundation models with new structural analysis perspectives, making Riemannian GFMs a promising future direction. For example, Riemannian manifolds play a role in constructing structural vocabularies and performing structural alignment among graphs. Ricci curvature tensors capture the diversity and commonality in the graph domain.
Identifying generalizable units for graphs is rather challenging in Euclidean space, but may be possible in Riemannian geometry using powerful tools such as diffeomorphisms and cohomology. 
Also, Riemannian graph learning is compatible with LLMs, and it is interesting to align them to construct the structural semantic space in Riemannian manifolds.

\subsection{Riemannian Graph Learning for Science (RG4S)}
As advances in artificial intelligence continue to reshape the landscape of scientific discovery, the paradigm of Artificial Intelligence for Science (AI4S) has emerged as a driving force across diverse disciplines, from quantum mechanics and molecular biology to climate modeling and astrophysics. 
Whether examining the curvature of spacetime in cosmology, deciphering the conformational dynamics of protein molecules, analyzing the topological patterns of brain networks, or unraveling the manifold structures of quantum states, the intimate connection between natural processes and geometric structures remains a foundational pillar of scientific exploration.

Riemannian graph learning has the potential to revolutionize scientific fields by providing powerful geometric tools. 
In life science,  the torus manifold has shown advantages in studying the $\alpha$-helix of proteins. 
In chemistry, SPD manifolds guarantee the $SO(n)$ equivariant, which is significant to molecular design.
In physics, Riemannian manifolds find their role in Hamiltonian mechanics, and thus underpins the exploration of physics-informed deep learning.
In Earth science, the geometric priors are important for solving partial differential equations, while graphs are often implicitly represented in neural operator learning. 
Riemannian geometry can benefit a wide spectrum of scientific disciplines, but it is largely ignored in recent advances in AI4S. Most deep learning models work with the Euclidean space and are unaware of the underlying structures and geometries. So far, there is no principled approach to activate Riemannian graph learning for science.

\section{Conclusion}  \label{sec: Conclusion}
In this survey, we provide a systematic and timely overview of Riemannian graph learning and offer a novel three-dimensional taxonomy that organizes this field along: (1) eight representative Riemannian manifolds, (2) six kinds of Riemannian neural architectures, and (3) four mainstream learning paradigms on manifolds. For each taxonomic category, we establish the mathematical frameworks, present a critical review of state-of-the-art methods, and provide comparative analysis and discussion. To facilitate empirical evaluation, we assemble benchmark datasets and open-source implementations. We further demonstrate the practical utility through representative applications that highlight the advantages of Riemannian approaches. The survey concludes by identifying key challenges and outlining promising research directions for next-generation Riemannian graph learning.


\bibliographystyle{IEEEtran}
\bibliography{IEEEtran}

\vskip -3.5\baselineskip
\clearpage

\end{document}